\documentclass[10pt,compsoc,twoside]{IEEEtran}  
\usepackage{epsfig}
\usepackage{graphicx}
\usepackage{amsmath}
\usepackage{amssymb}
\usepackage{algorithm}
\usepackage{algpseudocode}
\usepackage{multirow}
\usepackage{url}
\usepackage{tabularx}
\usepackage{color}
\usepackage[switch]{lineno}
\DeclareMathOperator*{\argmax}{argmax}
\newcommand{\eg}[1]{\textit{e.g.,}}
\newcommand{\ie}[1]{\textit{i.e.,}}
\newcommand{\etal}[1]{\textit{et al.}}
\newcommand{\etc}[1]{\textit{etc.}}

\ifCLASSOPTIONcompsoc
  \usepackage[nocompress]{cite}
\else
  \usepackage{cite}
\fi
\begin{document}
%
\title{Path-Restore: Learning Network Path \\ Selection for Image Restoration}
%
%
%
%

\author{Ke~Yu,
		Xintao~Wang,
		Chao~Dong, 
		Xiaoou~Tang,~\IEEEmembership{Fellow,~IEEE,}\\
		and~Chen~Change~Loy,~\IEEEmembership{Senior Member,~IEEE}
		\IEEEcompsocitemizethanks{
			\IEEEcompsocthanksitem K. Yu and X. Tang are with the Department of Information
			Engineering, The Chinese University of Hong Kong, Hong Kong. \hfil\break E-mail: \{yk017,xtang\}@ie.cuhk.edu.hk.
			
			\IEEEcompsocthanksitem X. Wang is with the Applied Research Center (ARC), Tencent PCG, Shenzhen, Guangdong 518054, China. E-mail: xintao.wang@outlook.com.
		
			\IEEEcompsocthanksitem{C. Dong is with the Guangdong-Hong Kong-Macao Joint Laboratory of Human-Machine Intelligence-Synergy Systems, Shenzhen Institute of Advanced Technology, Chinese Academy of Sciences, Shenzhen, Guangdong 518055, China. E-mail: chao.dong@siat.ac.cn.}
		
			\IEEEcompsocthanksitem{C. C. Loy (corresponding author) is with the S-Lab, Nanyang Technological University, Singapore 639798, Singapore. E-mail: ccloy@ntu.edu.sg.}
		}
    }
\markboth{IEEE Transactions on Pattern Analysis and Machine Intelligence}%
{Yu et al.: Path-Restore: Learning Network Path Selection for Image Restoration}
\IEEEtitleabstractindextext{%
\begin{abstract}
Very deep Convolutional Neural Networks (CNNs) have greatly improved the performance on various image restoration tasks. However, this comes at a price of increasing computational burden, hence limiting their practical usages. We observe that some corrupted image regions are inherently easier to restore than others since the distortion and content vary within an image. To leverage this, we propose Path-Restore, a multi-path CNN with a pathfinder that can dynamically select an appropriate route for each image region. We train the pathfinder using reinforcement learning with a difficulty-regulated reward. This reward is related to the performance, complexity and ``the difficulty of restoring a region''. 
A policy mask is further investigated to jointly process all the image regions.
We conduct experiments on denoising and mixed restoration tasks. The results show that our method achieves comparable or superior performance to existing approaches with less computational cost. In particular, Path-Restore is effective for real-world denoising, where the noise distribution varies across different regions on a single image. Compared to the state-of-the-art RIDNet~\cite{anwar2019real}, our method achieves comparable performance and runs 2.7x faster on the realistic Darmstadt Noise Dataset~\cite{plotz2017benchmarking}. Models and codes are available on the project page: \url{https://www.mmlab-ntu.com/project/pathrestore/}.
\end{abstract}

}

\maketitle

\IEEEdisplaynontitleabstractindextext

%
\IEEEpeerreviewmaketitle

\IEEEraisesectionheading{\section{Introduction}}
\IEEEPARstart{I}MAGE restoration aims at estimating a clean image from its distorted observation. Very deep Convolutional Neural Networks (CNNs) have achieved great success on various restoration tasks. For example, in the NTIRE 2018 Challenge~\cite{timofte2018ntire}, the top-ranking super-resolution methods are built on very deep models such as EDSR~\cite{lim2017enhanced} and DenseNet~\cite{huang2017densely}, achieving impressive performance even when the input images are corrupted with noise and blur. However, as the network becomes deeper and more complex, the increasing computational cost makes it less practical for real-world applications. 

\begin{figure}[t] \small
	\begin{center}
		\centering
		\includegraphics[width=\linewidth]{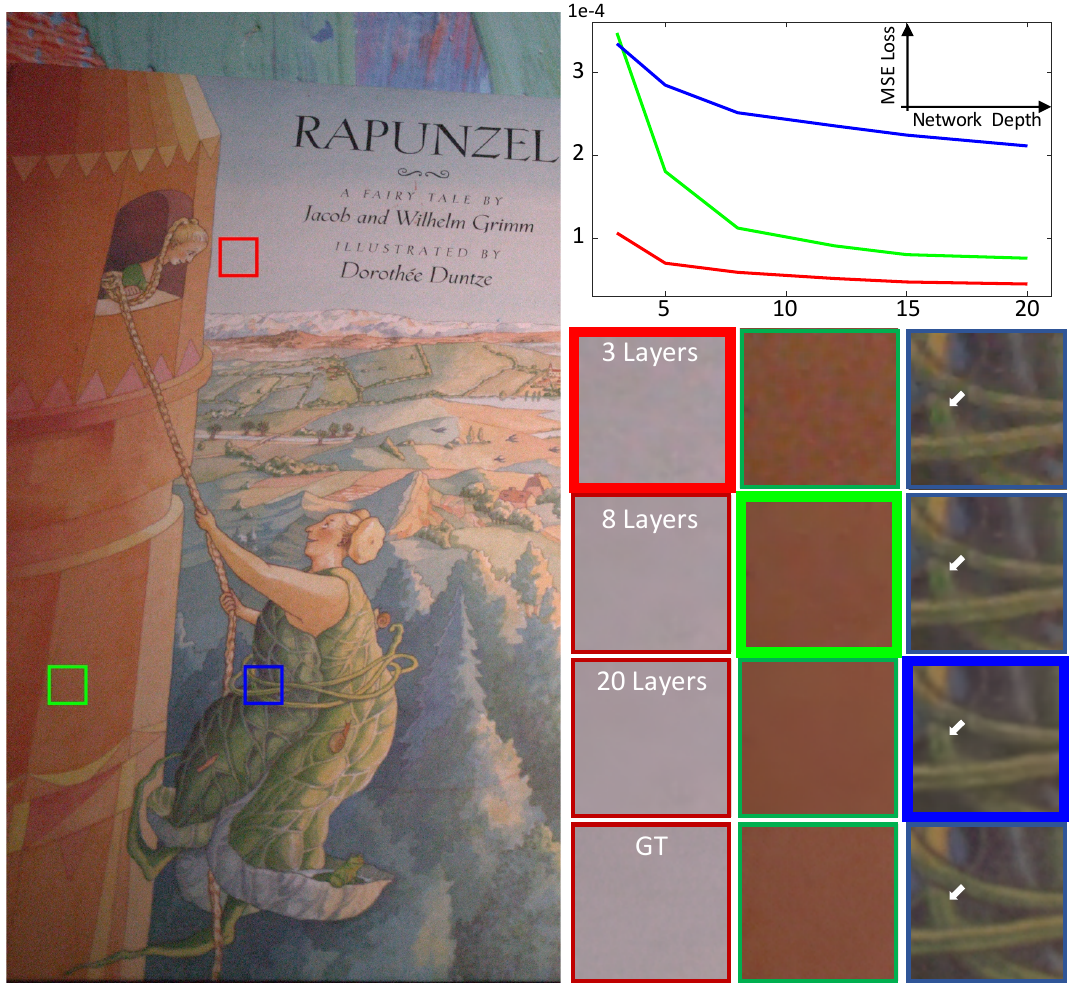}
		\caption{
			The relation between performance and network depth for different image regions. The noisy image is shown on the left. The restored outputs and the ground-truth image regions are presented on the right. Each curve represents the MSE loss of the corresponding region. The bold boxes denote a good trade-off between performance and complexity for each region. Zoom in for best view.
		}
		\label{fig:intro}
		\vspace{-0.5cm}
	\end{center}
\end{figure}

Do we really need a very deep CNN to process all the regions of a distorted image? In reality, a large variation of image content and distortion may exist in a single image, and some regions are inherently easier to process than others. An example is shown in Figure~\ref{fig:intro}, where we use several denoising CNNs with different depths to restore a noisy image. It is observed, in the red box, that a smooth region with mild noise is reasonably recovered with a 3-layer CNN; further increasing the network depth brings only incremental improvement. In the green box, a similar smooth region yet with severe noise requires an 8-layer CNN to process, whereas a texture region in the blue box still has some artifacts (highlighted by the white arrow) after being processed by a 20-layer CNN. This observation motivates the possibility of saving computations by selecting an optimal network path for each region based on its content and distortion.

In this study, we propose Path-Restore, a novel framework that can dynamically select a path for each image region with specific content and distortion. In particular, a pathfinder is trained together with a multi-path CNN to find the optimal dispatch policy for each image region. Since path selection is non-differentiable, the pathfinder is trained in a reinforcement learning (RL) framework driven by a reward that encourages both high performance and few computations. There are mainly two challenges. First, the multi-path CNN should be able to handle a large variety of distortions with limited computations. Second, the pathfinder requires an informative reward to learn a good policy for different regions with diverse contents and distortions.

\begin{figure*}[t] \small
	\centering
	\includegraphics[width=0.9\linewidth]{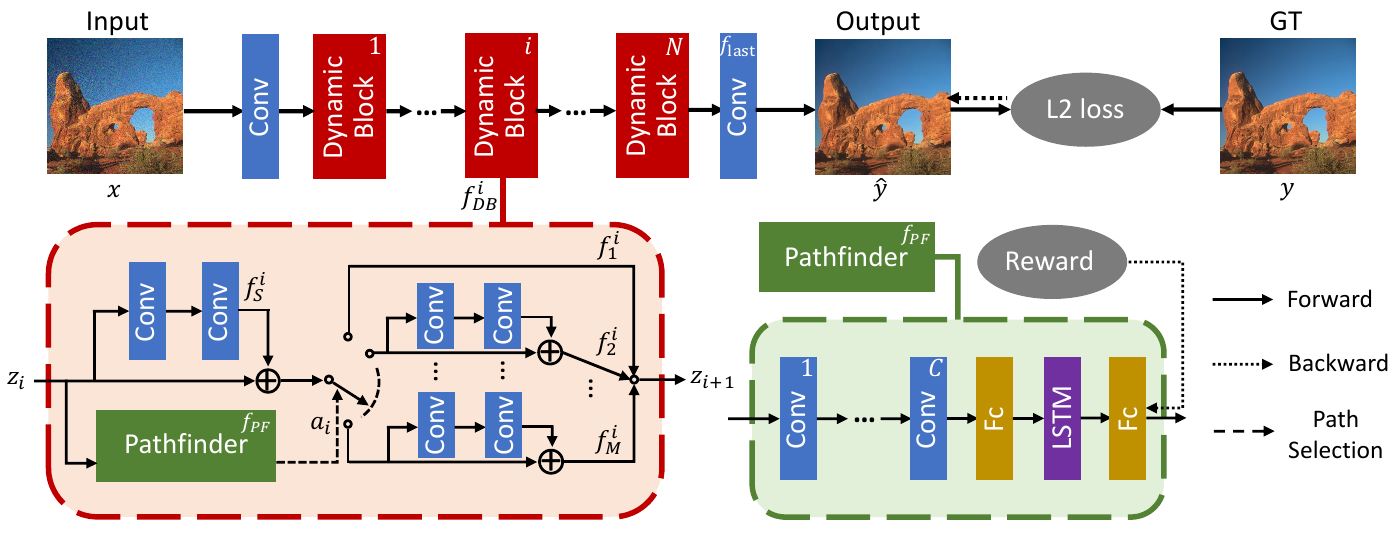}
	\caption{
		Framework Overview. Path-Restore is composed of a multi-path CNN and a pathfinder. The multi-path CNN contains $N$ dynamic blocks, each of which has $M$ optional paths. The number of paths is made proportional to the number of distortion types we aim to address. The pathfinder is able to dynamically select paths for different image regions according to their contents and distortions.
	}
	\label{fig:framework}
	\vspace{-0.3cm}
\end{figure*}

We make two contributions to address these challenges:

\noindent 
(1) We devise a dynamic block as the basic unit of our multi-path CNN. A dynamic block contains a shared path followed by several dynamic paths with different complexity. The proposed framework allows flexible design on the number of paths and path complexity according to the specific task at hand.

\noindent 
(2) We devise a difficulty-regulated reward. This reward values the performance gain of hard examples more than simple ones, unlike conventional reward functions. Specifically, the difficulty-regulated reward scales the performance gain of different regions by an adaptive coefficient that represents the difficulty of restoring this region. The difficulty is made proportional to a loss function (\eg, MSE loss), since a large loss suggests that the region is difficult to restore. For example, in Figure~\ref{fig:intro}, the loss of the green region decreases by a larger quantity than that of the blue region, resulting in a higher performance reward for the green region compared to the blue one. However, after regulated by a higher difficulty, the final reward of the blue region becomes larger, encouraging more computations for the restoration of hard regions. We experimentally find that this reward helps the pathfinder learn a more effective dispatch policy.

We further investigate two mechanisms to process different image regions. The first way is to treat all regions independently. During inference, an input image is first split into overlapping regions (patches of the same size). Then these regions are restored separately and merged together at last. This strategy is straightforward and easy to implement, thus is adopted in the basic version of Path-Restore. The second mechanism is to jointly process all regions at image level. The pathfinder first predicts a low-resolution policy mask, where each pixel represents a policy for a region in the input image. Then the features of different regions are sent to specific paths according to the corresponding policy on the predicted mask. With the policy mask, we can jointly process different regions with the guidance of adjacent features.
Our method with the second strategy, namely Path-Restore-Mask, can achieve a better performance-complexity trade-off.

We summarize the merits of Path-Restore as follows:
1) Thanks to the dynamic path-selection mechanism, our method achieves comparable or superior performance to existing approaches with faster speed on various restoration tasks. 
2) Our method is particularly effective in real-world denoising, where noise distribution is spatially variant within an image. We achieve state-of-the-art performance on the realistic Darmstadt Noise Dataset (DND)~\cite{plotz2017benchmarking} benchmark\footnote{\url{https://noise.visinf.tu-darmstadt.de/benchmark/#results_srgb}}.
3) Our method is capable of balancing the performance and complexity trade-off by adjusting a hyper-parameter in the proposed reward function during training.

\section{Related Work}

\noindent\textbf{Advances in Image Restoration.}
Image restoration has been extensively studied for decades. CNN-based methods have taken the leading role in several restoration tasks including denoising~\cite{lefkimmiatis2016non, jain2009natural, chen2018image, guo2019toward}, deblurring~\cite{xu2014Deep, sun2015learning, nah2017deep}, deblocking~\cite{dong2015compression, wang2016d3, guo2016building, guo2016one} and super-resolution~\cite{dong2016image, kim2016accurate, kim2016deeply, tai2017image, ledig2017photo, wang2018esrgan}.

Although most of the previous papers focus on addressing a specific degradation, several recent works aim at dealing with a large variety of distortions simultaneously. For example, Zhang~\etal~\cite{zhang2017beyond} propose DnCNN that employs a single CNN to address different levels of Gaussian noise. Guo~\etal~\cite{guo2019toward} develop CBDNet that estimates a noise map as a condition to handle diverse real noise. Plotz~\etal~\cite{plotz2018neural} devise a continuous and differentiable neural nearest neighbor block that works effectively on several restoration tasks. 
State-of-the-art denoising methods RIDNet~\cite{anwar2019real} and PRIDNet~\cite{zhao2019pyramid} adopt feature attention and pyramid architectures to enlarge network capacity, respectively. 
However, processing all image regions using a single path makes these methods less efficient. Smooth regions with mild distortions do not need such a deep network to restore.

The idea of region-wise processing has been explored in the literature. Anwar~\etal~\cite{anwar2017combined} propose a category-specific denoising algorithm. They apply internal denoising for smooth regions and category-specific external denoising for textured regions. 
We offer a thorough discussion about the differences between this method and our approach in Sec.~\ref{subsec:ablation}.

Our earlier work~\cite{yu2018crafting} proposes RL-Restore to address mixed distortions through applying deep reinforcement learning.
The method adaptively selects a sequence of CNNs to process each distorted image, achieving comparable performance with a single deep network while using fewer computational resources. Unlike RL-Restore that requires manual design for a dozen CNNs, we only use one network to achieve dynamic processing. Our method can thus be more easily migrated and generalized to various tasks. 
As Path-Restore does not need to switch among different models to process various regions, our method runs faster and produces more spatially consistent results than RL-Restore.
Moreover, our new framework enjoys the flexibility to balance the trade-off between performance and complexity by merely adjusting the reward function, a feature that is not possible in RL-Restore~\cite{yu2018crafting}.

\noindent\textbf{Dynamic Networks.}
Dynamic networks have been investigated to achieve a better trade-off between speed and performance in different tasks. 
Bengio~\etal~\cite{bengio2013estimating} use conditional computation to selectively activate one part of the network at a time. 
Recently, several approaches~\cite{figurnov2017spatially,wu2018blockdrop,wang2017skipnet} are proposed to save the computational cost in ResNet~\cite{he2016deep} by dynamically skipping some residual blocks. 
While the aforementioned methods mainly address one single task, Rosenbaum~\etal~\cite{rosenbaum2017routing} develop a routing network to facilitate multi-task learning, and their method seeks an appropriate network path for each specific task. However, the path selection is irrelevant to the image content in each task. 

Existing dynamic networks successfully explore a routing policy that depends on either the whole image content or the task to address. However, we aim at selecting dynamic paths to process different regions of a distorted image. In this case, both the content (\eg, smooth regions or rich textures) and the restoration task (\eg, distortion type and severity) have a weighty influence on the performance, and they should be both considered in the path selection. To address this specific challenge, we propose a dynamic block to offer diverse paths and adopt a difficulty-regulated reward to effectively train the pathfinder. The contributions are new in the literature.

\noindent\textbf{Handling Non-Differentiable Operations.} 
Deep reinforcement learning is comprehensively studied to address non-differentiable action selection. One of the most popular applications is decision making in games~\cite{mnih2015human,hessel2017rainbow,silver2017mastering}. A tremendous success is also achieved in neural architecture search~\cite{baker2016designing,zoph2016neural,zhong2018practical}, where the selection of a layer or block is not differentiable. Hu~\etal~\cite{hu2018exposure} develop an RL-based framework to learn a sequence of filters for image retouching.

Gumbel Softmax~\cite{jang2017categorical} is a widely-used technique to make a categorical distribution differentiable. For instance, Veit~\etal~\cite{veit2018convolutional} adopt Gumbel Softmax to achieve adaptive inference in a network. Xie~\etal~\cite{xie2019snas} and Wu~\etal~\cite{wu2019fbnet} leverage this technique for differentiable neural architecture search.
%
In this work, we use an RL-based approach to train the pathfinder, as RL offers more flexibility to devise the reward. The reward function is not necessarily differentiable. On the contrary, as a differentiable estimator, Gumbel Softmax requires a differentiable target function to optimize. Nevertheless, a differentiable estimator is worth trying in future work given its high training efficiency.

\section{Methodology}
\label{sec:method}

Our task is to recover a clear image $y$ from its distorted observation $x$.
We propose Path-Restore that can process each image region according to its content and distortion through a specific network path. We first provide an overview of the architecture of Path-Restore in Sec.~\ref{subsec:architecture}. In Sec.~\ref{subsec:pathfinder}, we then introduce the structure of pathfinder and the reward to evaluate its policy. 
In Sec.~\ref{subsec:path_restore_mask}, we describe Path-Restore-Mask that can jointly process all image regions.
Finally in Sec.~\ref{subsec:training}, we detail the training algorithm.

\begin{figure*}[t] \small
	\centering
	\includegraphics[width=0.9\linewidth]{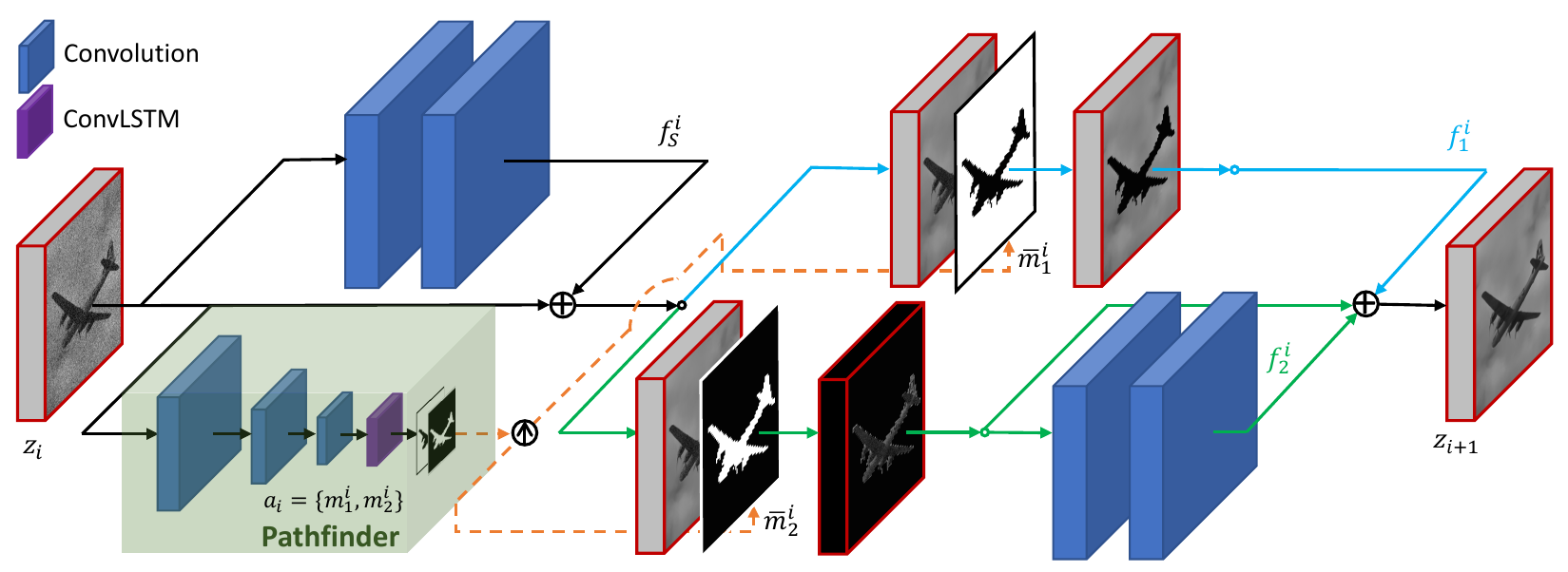}
	\caption{
		The architecture of dynamic block and pathfinder in Path-Restore-Mask. For simplicity, we show a special case with two paths. The pathfinder, depicted at the bottom left corner, contains three strided convolutions and a ConvLSTM~\cite{xingjian2015convolutional}. It predicts a policy mask with two channels, both of which are then up-sampled by nearest-neighbor interpolation to select eligible regions for the corresponding paths.
	}
	\label{fig:framework_mask}
	\vspace{-0.3cm}
\end{figure*}

\subsection{Architecture of Path-Restore}
\label{subsec:architecture}

We aim to design a framework that can offer different options of complexity.
To this end, as shown in Figure~\ref{fig:framework}, Path-Restore is composed of a convolutional layer at the start- and end-point, and $N$ dynamic blocks in the middle. In the $i$-th dynamic block $f_{DB}^i$, there is a shared path $f_S^i$ that every image region should pass through. Paralleling to the shared path, a pathfinder $f_{PF}$ generates a probabilistic distribution of plausible paths for selection. Following the shared path, there are $M$ dynamic paths denoted by $f_1^i, f_2^i, \dots, f_M^i$. Each dynamic path contains a residual block except that the first path is a bypass connection. According to the output of pathfinder, the dynamic path with the highest probability is activated, where the path index is denoted by $a_i$. For instance, if there are two dynamic paths in total, then $a_i\in\{1,2\}$. $a_i=1$ and $a_i=2$ represent that the first and the second path is chosen, respectively. The $i$-th dynamic block can be formulated as:
\begin{equation}
\label{eq:dynamic_block}
z_{i+1} = f_{a_i}^i(f_S^i(z_i)),
\end{equation}
where $z_i$ and $z_{i+1}$ denote the input and output of the \mbox{$i$-th} dynamic block, respectively.
Note that in two different dynamic blocks, the parameters of each corresponding path are different, while the parameters of pathfinder are shared. We use ReLU as the activation function between different layers. The last fully-connected layer of the pathfinder is activated by the Softmax function.

As can be seen from Eq.~\eqref{eq:dynamic_block}, the image features go through the shared path and one dynamic path. The shared path is designed to capture the similarity among different tasks and images. 
The dynamic paths offer different options of complexity. Intuitively, simple samples should be led to the bypass path to save computations, while hard samples might be guided to another path according to the content and distortion. 

\subsection{Pathfinder}
\label{subsec:pathfinder}

The pathfinder is the key component to achieve path selection. As shown in Figure~\ref{fig:framework}, the pathfinder contains $C$ convolutional layers, followed by two fully-connected layers with a Long Short-Term Memory (LSTM) module in the middle. The number of convolutional layers depends on the specific task that will be specified in Sec.~\ref{sec:exp}. The LSTM module is used to capture the correlations of path selection in different dynamic blocks. The pathfinder accounts for less than 3\% of the overall computations.

Since path selection is non-differentiable, we formulate the sequential path selection as a Markov Decision Process (MDP) and adopt reinforcement learning to train the pathfinder. We will first clarify the state and action of this MDP, and then illustrate the proposed difficulty-regulated reward to train the pathfinder.

\noindent\textbf{State and Action.} In the $i$-th dynamic block, the state $s_i$ consists of the input features $z_i$ and the hidden state of LSTM $h_i$, denoted by $s_i = \{z_i, h_i\}$. Given the state $s_i$, the pathfinder $f_{PF}$ generates a distribution of path selection that can be formulated as $f_{PF}(s_i) = \pi(a|s_i)$. In the training phase, the action (path index) is sampled from this probabilistic distribution, denoted by $a_i \sim \pi(a|s_i)$. While in the testing phase, the action is determined by the highest probability, \ie, $a_i = \argmax_a \pi(a|s_i)$.

\noindent\textbf{Difficulty-Regulated Reward.} In an RL framework, the pathfinder is trained to maximize a cumulative reward, and thus a proper design of reward function is critical. Existing dynamic models for classification~\cite{wu2018blockdrop,wang2017skipnet} usually use a trade-off between accuracy and computations as the reward. However, in our task, a more effective reward is required to learn a reasonable dispatch policy for different image regions that have diverse contents and distortions. Therefore, we propose a difficulty-regulated reward that not only considers performance and complexity, but also depends on ``the difficulty of restoring an image region''. In particular, the reward at the $i$-th dynamic block is formulated as:
\begin{equation}
\label{eq:reward}
{r_i=}\left\{
\begin{array}{ll}
-p\times(1-\mathbf{1}_{\{1\}}(a_i)), &1 \leq i < N,\\
-p\times(1-\mathbf{1}_{\{1\}}(a_i)) + d\times(-\Delta{L_2}), &i=N,
\end{array}
\right.
\end{equation}
where $-p$ is the reward penalty for choosing a complex path in one dynamic block. $\mathbf{1}_{\{1\}}(\cdot)$ represents an indicator function, \ie, $\mathbf{1}_{\{1\}}(a_i)=1$ only if $a_i=1$, otherwise $\mathbf{1}_{\{1\}}(a_i)=0$. $a_i$ is the selected path index in the $i$-th dynamic block. For example, $a_i=1$ denotes that the first path (bypass connection) is selected. In this case the reward is not penalized, \ie, the reward penalty term is multiplied by zero.
The performance and difficulty are only considered in the $N$-th dynamic block. In particular, $-\Delta{L_2}$ represents the performance gain in terms of L2 loss. The difficulty is denoted by $d$ in the following formula:
\begin{equation}
\label{eq:difficulty}
{d=}\left\{
\begin{array}{ll}
L_d / L_0, & 0 \leq L_d < L_0,\\
1, & L_d \geq L_0,
\end{array}
\right.
\end{equation}
where $L_d$ is the loss function of the output image and $L_0$ is a threshold. As $L_d$ approaches zero, the difficulty $d$ decreases, indicating the input region is easy to restore. In this case, the performance gain is regulated by $d < 1$ and the reward also becomes smaller, so that the proposed difficulty-regulated reward will penalize the pathfinder for wasting too much computation on easy regions. There are multiple choices for $L_d$ such as L2 loss and VGG loss~\cite{johnson2016perceptual}, and different loss functions may lead to different policy of path selection. 

\subsection{Path-Restore-Mask}
\label{subsec:path_restore_mask}

In Path-Restore, the pathfinder selects paths for each region without considering the selection results of adjacent regions. This may decrease the accuracy of path selection since the observed region is small. Moreover, as different regions are processed independently, adjacent regions must overlap to reduce the artifacts along borders. Overlapping regions will inevitably introduce additional computation cost. Therefore, we propose Path-Restore-Mask to process a high-resolution input at image level with higher efficiency.

The architecture of Path-Restore-Mask is illustrated in Figure~\ref{fig:framework_mask}. The overall structure is the same as Path-Restore (see Figure~\ref{fig:framework}), so we present the distinct parts -- dynamic block and the pathfinder in Figure~\ref{fig:framework_mask}.
Let $z_i$ and $z_{i+1}$ denote the input and output features of a large image, respectively. The shared path is denoted by $f_S^i$. For simplicity, we consider a special case with two dynamic paths ($M=2$), where the first path $f_1^i$ is a bypass connection and the second path $f_2^i$ is a residual block. The pathfinder is composed of three strided convolutional layers followed by a convolutional LSTM (ConvLSTM)~\cite{xingjian2015convolutional}. The action $a_i$ is a two-channel mask rather than a scalar. The first channel $m_1^i$ implies the regions that would go through the first path, and similarly the second mask corresponds to the second path. As depicted by the orange dash line, the two masks are then up-sampled by nearest-neighbor interpolation to the spatial resolution of $z_i$. Finally, the processed features of different paths are combined to generate the output features $z_{i+1}$. The whole process can be formulated as:
\begin{equation}
\label{eq:dynamic_block_mask}
z_{i+1} = \sum_{j=1}^{M}f_{j}^i(f_S^i(z_i) \odot \bar{m}_j^i),
\end{equation}
where $\bar{m}_j^i$ represents the up-sampled mask for the $j$-th path. The $\odot$ denotes element-wise multiplication, and each channel of features $f_S^i(z_i)$ performs the same element-wise multiplication with the mask $\bar{m}_j^i$.
Note that the regions with zero mask are ignored in each path, and hence the features of each region go through only one path.
We illustrate the architecture details and demonstrate the effectiveness of Path-Restore-Mask in Sec.~\ref{subsec:exp_mask}.


\begin{algorithm}[t]
	\caption{REINFORCE (1 update)}
	\label{alg:REINFORCE}
	\small
	\begin{algorithmic}
		\State Get a batch of $K$ image pairs, $\{x^{(1)}, y^{(1)}\}, \cdots, \{x^{(K)}, y^{(K)}\}$
		\State With policy $\pi$, derive $(s_1^{(k)}, a_1^{(k)}, r_1^{(k)}, \dots, s_N^{(k)}, a_N^{(k)}, r_N^{(k)})$
		\State Compute gradients using REINFORCE
		$$\Delta\theta = \frac{1}{K}\sum_{k=1}^{K}\sum_{i=1}^{N}\nabla_\theta\log\pi(a_i^{(k)}|s_i^{(k)};\theta)(\sum_{j=i}^{N}r_j^{(k)} - b^{(k)})$$
		Update parameters $\theta \gets \theta + \beta\Delta\theta$
	\end{algorithmic}
\end{algorithm}

\subsection{Training Algorithm}
\label{subsec:training}

The training process is composed of two stages. In the first stage, we train the multi-path CNN with random policy of path selection as a good initialization. In the second stage, we train the pathfinder and the multi-path CNN simultaneously, so that the two components are better associated and optimized.

\noindent\textbf{The First Stage.} We train the multi-path CNN with randomly selected routes. The path-selection policy is a uniform distribution if not specified. The loss function consists of two parts. First, a final L2 loss constrains the output images to have a reasonable quality. Second, an intermediate loss enforces the states observed by the pathfinder at each dynamic block to be consistent. Specifically, the loss function is formulated as:
\begin{equation}
\label{eq:loss_first_stage}
L = ||y-\hat{y}||_2^2 + \alpha\sum_{i=1}^{N}||y - f_{\text{last}}(z_i)||_2^2,
\end{equation}
where $\hat{y}$ and $y$ denote the output image and the ground truth, respectively. The $N$ represents the number of dynamic blocks. The last convolutional layer of Path-Restore is denoted by $f_{\text{last}}$, and $\alpha$ is the weight of the intermediate loss. If no intermediate loss is adopted, the features observed at different dynamic blocks may vary a lot from each other, hence making it more difficult for the pathfinder to learn a good dispatch policy.

\noindent\textbf{The Second Stage.} We train the pathfinder together with the multi-path CNN. In this stage, the multi-path CNN is trained using the loss function in Eq.~\eqref{eq:loss_first_stage} with $\alpha=0$, \ie, intermediate loss does not take effect. Meanwhile, we train the pathfinder using the REINFORCE algorithm~\cite{williams1992simple}, as shown in Algorithm~\ref{alg:REINFORCE}.
The parameters of pathfinder and learning rate are denoted by $\theta$ and $\beta$, respectively. We adopt an individual baseline reward $b^{(k)}$ to reduce the variance of gradients and stabilize training. In particular, $b^{(k)}$ is the reward of the $k$-th image when the bypass connection is selected in all dynamic blocks. 

\noindent\textbf{Implementation Details.}
As shown in Table~\ref{tab:settings}, the threshold of difficulty $L_0$ and the reward penalty $p$ are set based on the specific task. Generally it requires larger threshold and reward penalty when the distortions are more severe. For all tasks, the difficulty $d$ is measured by L2 loss. In the first training stage, the coefficient of intermediate loss $\alpha$ is set as 0.1. 
Both training stages take 800k iterations. Training images are cropped into 63$\times$63 patches and the batch size is 32. The learning rate is initially 2$\times10^{-4}$ and decayed by a half after every quarter of the training process. We use Adam~\cite{kingma2015adam} as the optimizer and implement our method on TensorFlow~\cite{abadi2016tensorflow}.

\begin{table}[t] \centering \small
	\center
	\caption{Settings of architecture and reward for different tasks.}
	\begin{tabular}{c|c|c|c|c|c}
		\hline
		\multirow{2}{*}{Task}  & \multicolumn{3}{c|}{Architecture}  & \multicolumn{2}{c}{Reward}    \\
		\cline{2-6}
		& $N$ & $M$ & $C$    & Threshold($L_0$)  & Penalty($p$) \\
		\hline\hline
		Denoising & 6   & 2   & 2      &  5$\times10^{-4}$ & 8$\times10^{-6}$ \\
		Mixed     & 5   & 4   & 4      &  0.01             & 4$\times10^{-5}$   \\
		\hline
		
	\end{tabular}
	\label{tab:settings}
\end{table}

\section{Experiments}
\label{sec:exp}
In this section, we first clarify the general settings of network architecture and evaluation details. In Sec.~\ref{subsec:real_noise}, we present the results of real-world denoising. In Sec.~\ref{subsec:denoising}, we demonstrate that our method performs well on blind Gaussian denoising. In Sec.~\ref{subsec:mixed}, we further show the capability of Path-Restore to address a mixture of complex distortions. In Sec.~\ref{subsec:exp_mask}, we verify the effectiveness of Path-Restore-Mask on real-world denoising. Finally, in Sec.~\ref{subsec:ablation}, we offer discussions about the architecture of dynamic block, the difficulty-regulated reward, the effects of pathfinder, the number of network parameters and more comparisons.

\noindent\textbf{Network Architecture.}
As shown in Table~\ref{tab:settings}, we select different architectures for different restoration tasks. Specifically, we employ $M=2$ (1 bypass, 1 denoising) paths for the denoising tasks while $M=4$ (1 bypass, 3 distortions) paths for addressing a mixed of three distortions. When the restoration task is more complex, we adopt more convolutional layers in the pathfinder. In particular, the pathfinder has $C=2$ and $C=4$ convolutional layers for denoising and addressing mixed distortions, respectively. As for the number of dynamic blocks, we use $N=6$ for denoising and $N=5$ for mixed distortions so that the network complexity is appropriate for a fair comparison with the baseline methods.

\begin{figure}[t] \small
	\centering
	\includegraphics[width=0.9\linewidth]{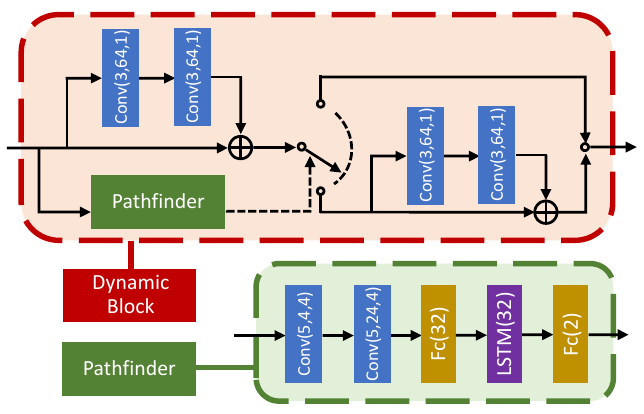}
	\caption{
		The architecture of dynamic block and pathfinder for all the denoising tasks.
	}
	\label{fig:architecture_denoise}
	\vspace{-0.3cm}
\end{figure}

\noindent\textbf{Evaluation Details.}
While testing Path-Restore, each image is split into 63$\times$63 regions with a stride 53. After being processed by the multi-path CNN, all the regions are merged into a large image with overlapping pixels averaged. Following RL-Restore~\cite{yu2018crafting}, we report the number of floating point operations (FLOPs) that are required to process a 63$\times$63 region on average. Each multiplication or addition contributes one FLOP. We also record the GPU\footnote{NVIDIA GeForce GTX TITAN X} or CPU\footnote{Intel(R) Xeon(R) CPU E5-2640 v3 @ 2.60GHz} runtime. We report the time to process an image of size 512$\times$512 if not specified.

\subsection{Evaluation on Real-World Denoising}
\label{subsec:real_noise}

\noindent\textbf{Architecture Details.}
We first conduct experiments on real-word denoising. The detailed architectures of dynamic block and the pathfinder are shown in Figure~\ref{fig:architecture_denoise}. Conv($k$,$n$,$m$) denotes a convolutional layer with a $k\times{k}$ kernel size, $n$ filters and a stride of $m$. A fully-connected layer with output size $n_o$ and an LSTM module with hidden size $n_h$ are denoted by Fc($n_o$) and LSTM($n_h$), respectively. 

\noindent\textbf{Training and Testing Details.}
We evaluate our method on the Darmstadt Noise Dataset (DND)~\cite{plotz2017benchmarking} and the Smartphone Image Denoising Dataset (SIDD)~\cite{SIDD_2018_CVPR}. The DND contains 50 realistic high-resolution paired noisy and noise-free images for evaluation, yet it does not offer data for training. For DND benchmark, we adopt the same training data as CBDNet~\cite{guo2019toward} for a fair comparison. In particular, we use BSD500~\cite{martin2001database} and Waterloo~\cite{ma2017waterloo} datasets to synthesize noisy images, and use the same noise model as CBDNet. A real dataset RENOIR~\cite{anaya2018renoir} is also adopted for training as CBDNet. 
As for SIDD benchmark, we use SIDD Medium dataset for training.
For non-blind denoising methods like BM3D~\cite{dabov2007color} and FFDNet~\cite{zhang2018ffdnet}, we follow~\cite{zhao2019pyramid} to exploit Neat Image, a plug-in of Photoshop, for noise estimation.

\begin{table}[t] \centering \small
	\center
	\caption{Results of real-world denoising on the Darmstadt Noise Dataset~\cite{plotz2017benchmarking}. The ``$\sim$'' represents an estimation of FLOPs. The time is tested on the same GPU for a 512$\times$512 image.}
	\begin{tabular}{c|c|c|c}
		\hline
		Method                        & PSNR / SSIM                      & FLOPs (G)     &  Time (s)                \\ \hline\hline
		BM3D~\cite{dabov2007color}    & 34.51 / 0.8507                   & -             &  11                      \\
		FFDNet~\cite{zhang2018ffdnet} & 37.61 / 0.9415                   & 1.74          &  0.051                   \\
		CBDNet~\cite{guo2019toward}   & 38.06 / 0.9421                   & 6.94          &  0.19                    \\
		N3Net~\cite{plotz2018neural}  & 38.32 / 0.9384                   & $\sim$15.4         &  0.67                    \\
		RIDNet~\cite{anwar2019real}   & 39.26 / 0.9528                   & 13.0          &  0.41                    \\
		PRIDNet~\cite{zhao2019pyramid}& \underline{39.42} / 0.9528       & 3.65          &  0.077                   \\
		\hline
		Path-Restore                & 39.00 / \underline{0.9542}         & 5.60          &  0.15                    \\
		Path-Restore-Ext            & \textbf{39.72} / \textbf{0.9591}   & 22.6          &  0.42                    \\ \hline
	\end{tabular}
	\label{tab:DND}
	\vspace{-0.2cm}
\end{table}

\begin{table}[t] \centering \small
	\center
	\caption{Results of real-world denoising on the SIDD Dataset~\cite{SIDD_2018_CVPR}. The runtime is directly copied from the benchmark for reference.}
	\begin{tabular}{c|c|c|c}
		\hline
		Method                        &\ PSNR\  &\ SSIM\                        &  Time (s/Mpixel)     \\ \hline\hline
		BM3D~\cite{dabov2007color}    & 25.65 & 0.685                           &  27.4                \\
		CBDNet~\cite{guo2019toward}   & 33.28 & 0.868                           &  4.48                \\
		Proxy-opt. BM3D~\cite{tseng2019hyperparameter} & 34.34  & 0.911         &  6.69                \\
		Path-Restore                  & \textbf{38.21} & \textbf{0.946}         &  \textbf{0.89}       \\
		\hline
	\end{tabular}
	\label{tab:SIDD}
	\vspace{-0.2cm}
\end{table}

\begin{figure*}[h] \small
	\centering
	\includegraphics[width=0.95\linewidth]{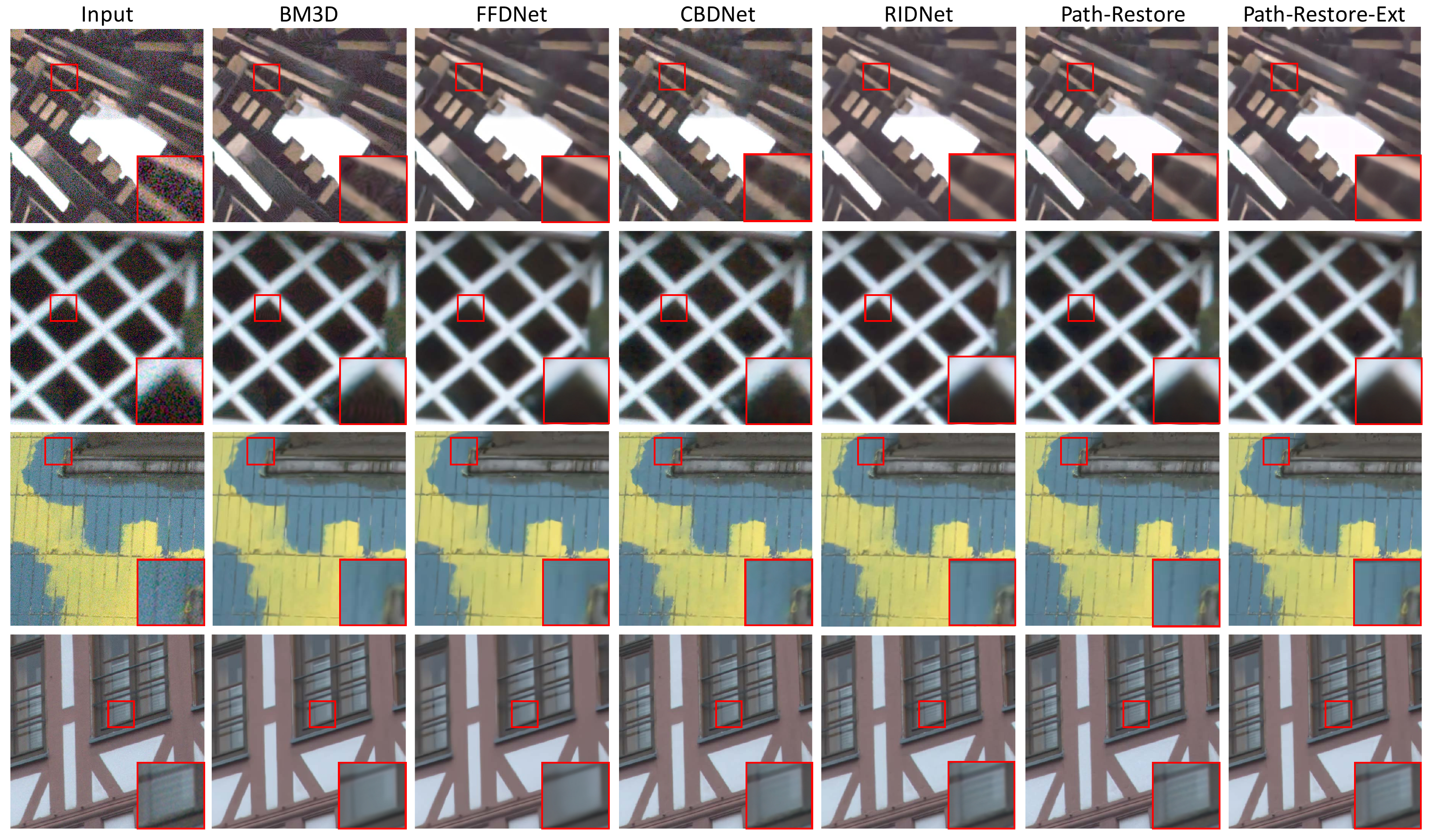}
	\vspace{-0.3cm}
	\caption{
		Qualitative results on the Darmstadt Noise Dataset~\cite{plotz2017benchmarking}. Path-Restore recovers clean results with sharp edges.
	}
	\label{fig:image_CBD}
	
	\vspace{0.2cm}
	\centering
	\includegraphics[width=0.95\linewidth]{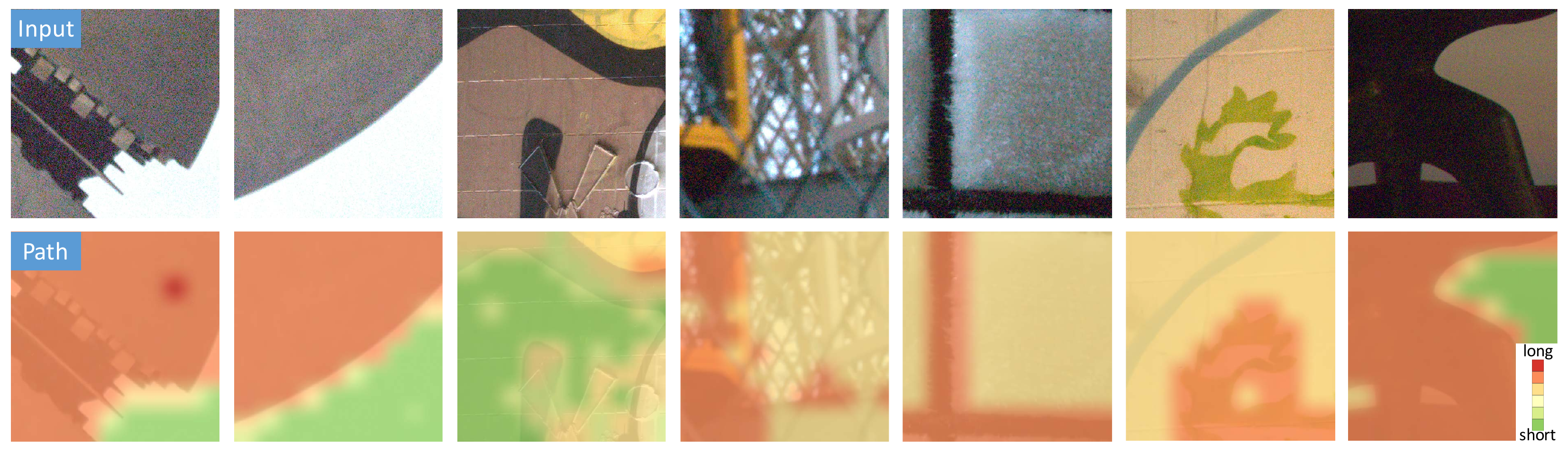}
	\vspace{-0.3cm}
	\caption{
		The policy of path selection on DND~\cite{plotz2017benchmarking}. The green color represents a short path while the red color represents a long path. Dark regions with severe noise are processed more than bright regions with slight noise.
	}
	\label{fig:heatmap_CBD}
\end{figure*}

\begin{table*}[t] \centering \small
	\center
	\caption{PSNR and average FLOPs of blind Gaussian denoising on CBSD68 and DIV2K-T50 datasets. The unit of FLOPs is Giga ($\times10^9$). Path-Restore is consistently 25\% faster (in terms of FLOPs) than DnCNN with comparable performance on different noise settings.}
	\begin{tabular}{c|c|c|c|c|c|c|c|c|c|c|c|c}
		\hline
		Dataset                      & \multicolumn{6}{c|}{CBSD68~\cite{roth2009fields}}                                   & \multicolumn{6}{c}{DIV2K-T50~\cite{agustsson2017ntire}}                 \\ \hline
		\multirow{2}{*}{Noise}       & \multicolumn{3}{c|}{uniform}               & \multicolumn{3}{c|}{spatially variant} & \multicolumn{3}{c|}{uniform}               & \multicolumn{3}{c}{spatially variant}  \\ \cline{2-13} 
		& $\sigma$=10 & $\sigma$=50 & FLOPs          & linear    & peaks   & FLOPs            & $\sigma$=10 & $\sigma$=50 & FLOPs          & linear    & peaks   & FLOPs            \\ \hline\hline
		DnCNN~\cite{zhang2017beyond} & 36.07       & 27.96       & 5.31G          & 31.17     & 31.15   & 5.31G            & 37.32       & 29.64       & 5.31G          & 32.82     & 32.64   & 5.31G            \\
		Path-Restore                 & 36.04       & 27.96       & \textbf{4.22G} & 31.18     & 31.15   & \textbf{4.22G}   & 37.26       & 29.64       & \textbf{4.20G} & 32.83     & 32.64   & \textbf{4.17G}   \\ \hline
	\end{tabular}
	\label{tab:denoising_results}
\end{table*}

\begin{figure*}[t] \small
	\centering
	\includegraphics[width=0.94\linewidth]{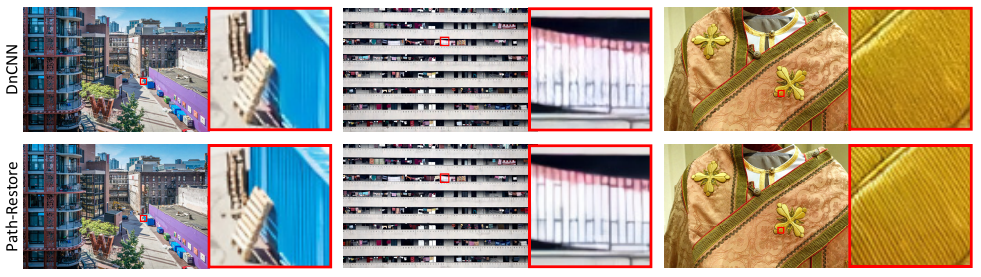}
	\vspace{-0.2cm}
	\caption{
		Qualitative results of spatially variant (type ``peaks'') Gaussian denoising. While most visual results are comparable, Path-Restore recovers textured regions better than DnCNN since these regions are processed with long paths.
	}
	\label{fig:image_denoise}
	\vspace{0.1cm}
	
	\centering
	\includegraphics[width=0.95\linewidth]{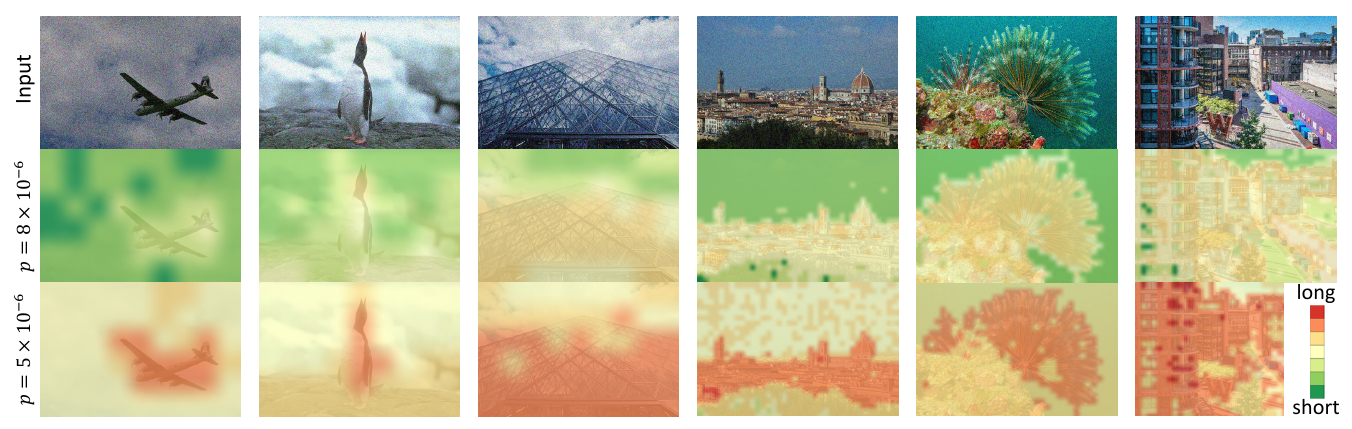}
	\vspace{-0.2cm}
	\caption{
		The policy of path selection for uniform Gaussian denoising $\sigma=30$. The pathfinder learns to select long paths for the objects with detailed textures while process the smooth background with short paths.
	}
	\label{fig:heatmap_denoise}
	\vspace{-0.3cm}
\end{figure*}
\begin{figure*}[h]
	\vspace{0.3cm}
	
	\centering
	\includegraphics[width=0.95\linewidth]{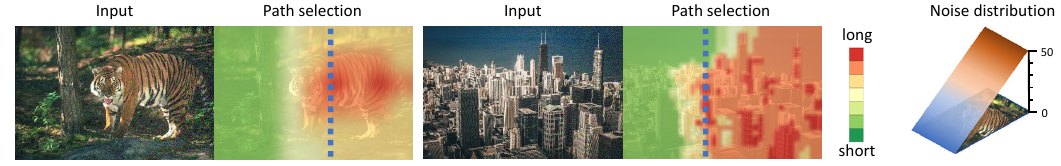}
	\caption{
		The policy of path selection for spatially variant (type ``linear'') Gaussian denoising. The pathfinder learns a dispatch policy based on both the content and the distortion, \ie, short paths for smooth and clean regions while long paths for textured and noisy regions.
	}
	\label{fig:heatmap_denoise_sv}
\end{figure*}

\noindent\textbf{Quantitative Results.}
The quantitative results on DND~\cite{plotz2018neural} are presented in Table~\ref{tab:DND}. Besides PSNR and SSIM metrics, we also report the FLOPs and GPU runtime. Path-Restore is significantly better than the non-blind denoising methods BM3D~\cite{dabov2007color} and FFDNet~\cite{zhang2018ffdnet}. 
Using the same training dataset as CBDNet~\cite{guo2019toward}, Path-Restore improves the performance by nearly 1 dB with 19\% fewer FLOPs and 27\% faster speed on GPU. Compared with N3Net~\cite{plotz2017benchmarking} that applies denoising to the raw image, Path-Restore attains better performance and runs faster. 
Our method is comparable to RIDNet~\cite{anwar2019real} with only about 40\% computation and runtime. PRIDNet~\cite{zhao2019pyramid} has higher PSNR and runs faster than our method, but its number of parameters is larger than ours by two orders of magnitude (74.2 million for PRIDNet and 0.9 million for ours).
Path-Restore-Ext denotes an extended deeper model trained with more data. This extended model further improves 0.7 dB and currently achieves the state-of-the-art performance on the DND benchmark. The details are specified at the end of this section.

The quantitative comparisons on SIDD are shown in Table~\ref{tab:SIDD}. It is observed that Path-Restore surpasses the state-of-the-art proxy-optimized BM3D~\cite{dabov2007color} by a large margin. We report the GPU runtime of Path-Restore for a 1024$\times$1024 input image (1 Mega pixels). For other methods, we directly copy the time recorded on the SIDD benchmark\footnote{\url{https://www.eecs.yorku.ca/~kamel/sidd/benchmark.php}}. Path-Restore is the most efficient method among all the existing works on the SIDD benchmark.

\noindent\textbf{Qualitative Results.}
Qualitative results are shown in Figure~\ref{fig:image_CBD}. We do not present the results of N3Net~\cite{plotz2018neural} because it is applied in the raw domain. 
In Figure~\ref{fig:image_CBD}, the top two input images are corrupted with severe noise, while the bottom two have moderate noise. It is observed that BM3D~\cite{dabov2007color} introduces artifacts for severe noise and yields over-smooth results for mild noise. FFDNet~\cite{zhang2018ffdnet} could successfully remove severe noise, but the results are too smooth and some tiny edges are almost missing.
CBDNet~\cite{guo2019toward} fails to remove the severe noise in dim regions and tends to generate artifacts along edges. This may be caused by the incorrect noise estimation for real-world images. 
The results of RIDNet are comparable to ours. RIDNet~\cite{anwar2019real} produces fewer artifacts while our method generates sharper edges.
With a deeper architecture and more training data, Path-Restore-Ext further mitigates the artifacts along sharp edges (\eg, the highlighted region of the first image).

\noindent\textbf{Path Selection.}
We present the policy of path selection in Figure~\ref{fig:heatmap_CBD}. The green color represents a short and simple path while the red color stands for a long and complex path. Path-Restore develops a reasonable policy for real-world denoising, using longer paths to process dark regions with severe noise while using shorter paths to process brighter regions with slight noise.

\noindent\textbf{Path-Restore-Ext.}
To train the extended Path-Restore-Ext, we 1) adopt a deeper network, 2) use more training data, 3) reduce the reward penalty and 4) enlarge the patch size for training and testing. In particular, the network architecture is two times deeper than Path-Restore. We use more training data from the SIDD Small dataset~\cite{SIDD_2018_CVPR}. Inspired by~\cite{brooks2018unprocessing}, we use unprocessing to get clear raw images with a similar distribution to those in DND~\cite{plotz2017benchmarking}. Following~\cite{brooks2018unprocessing}, we add synthetic noise to the raw images. We then do raw processing for both clean and noisy raw images to get a pair of training images in sRGB domain. These images are mixed with the training data of Path-Restore, providing a larger dataset to train Path-Restore-Ext. We further reduce the reward penalty $p$ from $8\times10^{-6}$ to $2\times10^{-7}$ in order to encourage longer paths and better performance. The training and testing patch size are increased from 63$\times$63 to 96$\times$96. While testing, the stride between adjacent patches is 80.

\subsection{Evaluation on Blind Gaussian Denoising}
\label{subsec:denoising}

\noindent\textbf{Training and Testing Details.}
We further conduct experiments on a blind denoising task where the distortion is Gaussian noise with a wide range of standard deviation [0, 50]. The network architecture is the same as that for real-world denoising. The training dataset includes the first 750 images of DIV2K~\cite{agustsson2017ntire} training set (denoted by DIV2K-T750) and 400 images used in~\cite{chen2017trainable}. We choose DnCNN~\cite{zhang2017beyond} as our baseline since it has similar number of parameters as our method (0.7 million for DnCNN and 0.9 million in total for ours). For fair comparison, we use the officially released codes to train DnCNN~\cite{zhang2017beyond} with the same training dataset as ours. Following~\cite{zhang2017beyond}, we test our model on CBSD68\footnote{Color images of BSD68 dataset~\cite{roth2009fields}.} and we further do evaluation on the remaining 50 images in the DIV2K training set (denoted by DIV2K-T50). In addition to uniform Gaussian noise, we also conduct experiments on spatially variant Gaussian noise using the settings in FFDNet~\cite{zhang2018ffdnet}. 
Specifically, ``linear'' denotes that the Gaussian noise level gradually increases from 0 to 50 from the left to the right side of an image. Another spatially variant noise, denoted by ``peaks'', is obtained by translating and scaling Gaussian distributions as in~\cite{zhang2018ffdnet}, and the noise level also ranges from 0 to 50.

\noindent\textbf{Quantitative and Qualitative Results.}
We report the PSNR performance and average FLOPs in Table~\ref{tab:denoising_results}. For both uniform and spatially variant denoising on different datasets, Path-Restore is able to achieve about 25\% speed-up (in terms of FLOPs) than DnCNN~\cite{zhang2017beyond} while achieving comparable performance. The acceleration on DIV2K-T50 is slightly larger than that on CBSD68, because high-resolution images in DIV2K-T50 tend to have more smooth regions that can be restored in short network paths.
%
When processing a 512$\times$512 image, the CPU runtime of DnCNN is 3.16s, while the runtime of Path-Restore is 2.37s and 2.99s given $\sigma=10$ and $\sigma=50$, respectively. In practice, Path-Restore is about 20\% faster than DnCNN on average.

Several qualitative results of spatially variant (type ``peaks'') Gaussian denoising are shown in Figure~\ref{fig:image_denoise}. Although most of the restored images are visually comparable, Path-Restore achieves better results on some regions with fine-grained textures, since Path-Restore could select more complex paths to process hard examples with detailed textures.

\noindent\textbf{Path Selection.}
We first visualize the path selection for uniform Gaussian denoising in Figure~\ref{fig:heatmap_denoise}. The input noisy images are shown in the first row, and the policy under $p=8\times10^{-6}$ and $p=5\times10^{-6}$ are presented in the second and third rows, respectively.  

It is observed that a larger reward penalty $p$ leads to a shorter path, as the pathfinder learns to avoid choosing overly complex paths that have very high reward penalty. Although the noise level is uniform within each image, the pathfinder learns a dispatch policy that depends on the image content. For example, the pathfinder focuses on processing the subject of an image (\eg, the airplane and penguin in the left two images) while assigning the smooth background to simple paths.

\begin{figure}[t] \small
	\vspace{-0.1cm}
	\centering
	\includegraphics[width=0.9\linewidth]{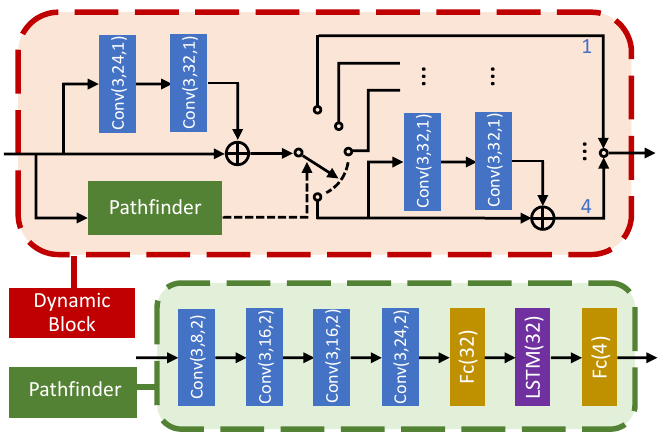}
	\caption{
		The architecture of dynamic block and pathfinder for addressing mixed distortions.
	}
	\label{fig:architecture_mix}
	\vspace{-0.1cm}
\end{figure}

The dispatch policy for spatially variant (type ``linear'') Gaussian denoising is shown in Figure~\ref{fig:heatmap_denoise_sv}. As the noise level gradually increases from left to right, the chosen paths become more and more complex. This demonstrates that the pathfinder learns a dispatch policy based on distortion level. Moreover, on each blue dash line, the noise level remains the same but the path selection is quite diverse, again demonstrating the pathfinder could select short paths for smooth regions while choose long paths for textured regions.

\begin{figure*}[t] \small
	\vspace{-0.1cm}
	\centering
	\includegraphics[width=0.95\linewidth]{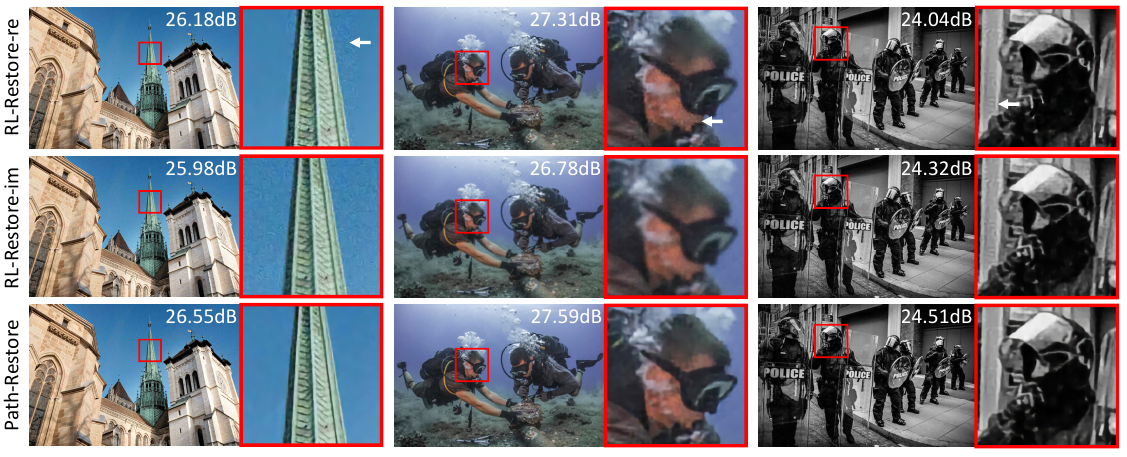}
	\caption{
		Qualitative results of addressing mixed distortions. RL-Restore~\cite{yu2018crafting} tends to generate artifacts across different regions (see the white arrow), while Path-Restore is able to handle diverse distortions and produce more spatially consistent results (zoom in for best view).
	}
	\label{fig:image_rl-restore}
	\vspace{-0.2cm}
\end{figure*}

\begin{figure*}[t] \small
	\centering
	\includegraphics[width=0.95\linewidth]{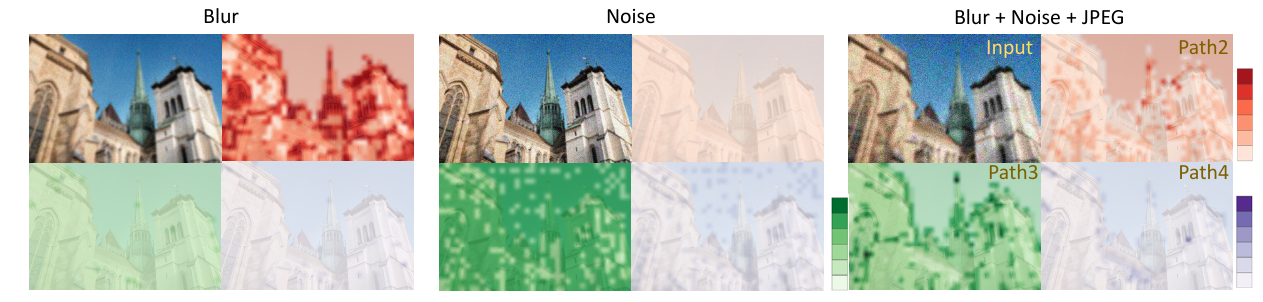}
	\caption{
		The policy of path selection for mixed distortions (4 paths including 1 bypass connection in each dynamic block). The light color represents that a small number of the corresponding paths are selected, while the dark color shows that a large number of the corresponding paths are chosen.
	}
	\label{fig:heatmap_mix}
\end{figure*}

\subsection{Evaluation on Mixed Distortions}
\label{subsec:mixed}

\noindent\textbf{Training and Testing Details.}
We further evaluate our method on a complex restoration task as that in RL-Restore~\cite{yu2018crafting}, where an image is corrupted by different levels of Gaussian blur, Gaussian noise and JPEG compression simultaneously. 
The network architecture is presented in Figure~\ref{fig:architecture_mix}, where there are four paths in each dynamic block.
Following~\cite{yu2018crafting}, we use DIV2K-T750 for training and DIV2K-T50 for testing. We report results not only on 63$\times$63 sub-images as in~\cite{yu2018crafting} but also on the whole images with 2K resolution. 
To conduct a thorough comparison, we test RL-Restore in two ways for large images: 1) each 63$\times$63 region is processed using a specific toolchain (denoted by RL-Restore-re), and 2) each 63$\times$63 region votes for a toolchain, and the whole image is processed with a unified toolchain that wins the most votes (denoted by RL-Restore-im).

\begin{table}[t] \centering \small
	\center
	\caption{Results of addressing mixed distortions on DIV2K-T50~\cite{agustsson2017ntire} compared with two variants of RL-Restore~\cite{yu2018crafting}.}
	\begin{tabular}{c|c|l|c|l|c}
		\hline
		\hspace{-0.1cm}Image size\hspace{-0.1cm}        & \multicolumn{2}{c|}{63x63 image}             & \multicolumn{3}{c}{2K image}                                    \\ \hline
		\hspace{-0.1cm}Metric\hspace{-0.1cm}            & \multicolumn{2}{c|}{PSNR / SSIM}             & \multicolumn{2}{c|}{PSNR / SSIM}             & \hspace{-0.1cm}FLOPs(G)\hspace{-0.1cm}         \\ \hline\hline
		\hspace{-0.1cm}RL-Restore-re\hspace{-0.1cm}         & \multicolumn{2}{c|}{-}                        & \multicolumn{2}{c|}{25.61 / 0.8264}          & \hspace{-0.1cm}1.34 \hspace{-0.1cm}             \\
		\hspace{-0.1cm}RL-Restore-im\hspace{-0.1cm}          & \multicolumn{2}{c|}{26.45 / 0.5587}          & \multicolumn{2}{c|}{25.55 / 0.8251}          & \hspace{-0.1cm}\textbf{0.948}\hspace{-0.1cm}    \\
		\hspace{-0.1cm}Path-Restore\hspace{-0.1cm}     & \multicolumn{2}{c|}{\textbf{26.48 / 0.5667}} & \multicolumn{2}{c|}{\textbf{25.81 / 0.8327}} & \hspace{-0.1cm}1.38\hspace{-0.1cm}              \\ \hline
	\end{tabular}
	\label{tab:mix_results}
	\vspace{-0.2cm}
\end{table}

\begin{table*}[t] \centering \small
	\center
	\caption{Quantitative evaluation of Path-Restore-Mask on the Darmstadt Noise Dataset~\cite{plotz2017benchmarking} and the Smartphone Image Denoising Dataset~\cite{SIDD_2018_CVPR}.}
	\setlength{\tabcolsep}{2.3mm}
	\begin{tabular}{c|c|c|c|c|c|c|c|c}
		\hline
		Dataset & \multicolumn{4}{c|}{DND~\cite{plotz2017benchmarking}} & \multicolumn{4}{c}{SIDD~\cite{SIDD_2018_CVPR}} \\
		\hline
		Metric                      & \ PSNR\        &\ \ SSIM\ \       & FLOPs (G)     & Time (s)            & \ PSNR\          &\ \ SSIM\ \     & FLOPs (G)      & Time (s/Mpixel)  \\ 
		\hline\hline
		Path-Restore                & 39.00          & 0.9542           & 5.60          & 0.149               & 38.21            & 0.946          & 7.13           & 0.89             \\
		Path-Restore-Mask           & \textbf{39.18} & \textbf{0.9563}  & \textbf{4.77} & \textbf{0.146}      & \textbf{38.42}   & \textbf{0.949} & \textbf{6.34}  & \textbf{0.76}    \\ 
		\hline
	\end{tabular}
	\label{tab:dnd_sidd_mask}
	\vspace{-0.2cm}
\end{table*}

In the first training stage, the random policy (initialization of path section) is a categorical distribution related to the specific combination of distortions. In particular, we classify each type of distortion (\ie, Gaussian blur, Gaussian noise and JPEG compression) into 10 levels (from 1 to 10) in the same way as RL-Restore~\cite{yu2018crafting}. The degradation levels of blur, noise and JPEG are denoted by $l_b$, $l_n$ and $l_j$, respectively. The parameters of categorical distribution can be written as $\sigma(\boldsymbol{l})$, where $\sigma(\cdot)$ is the Softmax function and $\boldsymbol{l}=(1, 0.2l_b-0.1, 0.2l_n-0.1, 0.2l_j-0.1)$. Given this prior distribution, if an image is mainly corrupted by one type of distortion, it will be assigned to the corresponding path with a high probability. If none of these three types of distortion is severe, the input image will be dispatched to the first path (bypass connection). 

\noindent\textbf{Quantitative Results.}
The quantitative results are shown in Table~\ref{tab:mix_results}. The results on 63$\times$63 sub-images show that Path-Restore achieves a slightly better performance compared with RL-Restore. While testing on large images with 2K resolution, our method shows more apparent gain (0.2 dB higher)  than RL-Restore-re and RL-Restore-im with just a tiny increase in computational complexity (FLOPs). The CPU runtime of RL-Restore-re, RL-Restore-im and Path-Restore are 1.34s, 1.09s and 0.954s, respectively. These results show that Path-Restore runs faster than RL-Restore in practice. The reason is that RL-Restore needs extra time to switch among different models while this overhead is not needed in the unified framework -- Path-Restore.

\noindent\textbf{Qualitative Results.}
As shown in Figure~\ref{fig:image_rl-restore}, RL-Restore-re tends to produce inconsistent boundaries and appearance between adjacent 63$\times$63 regions, as each region may be processed by entirely different models. RL-Restore-im fails to address severe noise or blur in some regions because only a single toolchain can be selected for the whole image. On the contrary, thanks to the dynamic blocks in a unified network, Path-Restore not only removes the distortions in different regions but also yields spatially consistent results.

\noindent\textbf{Path Selection.}
We show the results of path selection in Figure~\ref{fig:heatmap_mix}. The number of the 2nd, 3rd and 4th path selected are shown in the red, green and purple heat maps, respectively. For a blurry image with slight noise and compression, the pathfinder dispatches almost all image regions to the 2nd path, namely, more blurry images are assigned to the 2nd path compared with other paths, during the first training stage. Smooth regions are routed to the bypass path even when the blur is severe, because blur (without other types of distortion) does not affect the looking of smooth regions. Given an image with severe noise, the pathfinder routes most regions to the 3rd path, and even smooth regions require long paths to restore when the noise level is high. When multiple distortions co-exist, the path selection depends on both the image content and the combination of distortions.
Interestingly, the 4th path that has been assigned many JPEG images in the first training stage, is frequently selected at the first dynamic block yet seldom chosen at successive blocks. Perhaps the pathfinder learns that addressing JPEG compression at the beginning is a good restoration policy.

\subsection{Evaluation of Path-Restore-Mask}
\label{subsec:exp_mask}

\noindent\textbf{Implementation Details.}
In Path-Restore-Mask, the architectures of the shared path and dynamic paths are the same as those of Path-Restore (in Sec.~\ref{subsec:real_noise}). As for the pathfinder, the convolutional layers are ``Conv(5,16,4)--Conv(5,8,4)--Conv(3,2,2)'', where the parameters denote kernel size, number of filters and stride, respectively. The hidden state of ConvLSTM has two channels. Note that one pixel on the mask represents a 32$\times$32 region on the features. 
The training data is the same as Path-Restore, as specified in Sec.~\ref{subsec:real_noise}. We evaluate the performance on DND~\cite{plotz2017benchmarking} and SIDD~\cite{SIDD_2018_CVPR} benchmarks for real-world denoising.

\noindent\textbf{Quantitative Results.}
Quantitative comparisons between Path-Restore-Mask and Path-Restore are shown in Table~\ref{tab:dnd_sidd_mask}. On both DND and SIDD benchmarks, Path-Restore-Mask consistently outperforms Path-Restore by more than 0.2 dB. The better performance does not come at a price of higher complexity. It is observed that the FLOPs of Path-Restore-Mask are even 10\% fewer than Path-Restore on both benchmarks, and the GPU runtime of Path-Restore-Mask is also shorter. Processing a large image at image level is more efficient than at patch level. When treating different regions jointly, the features of adjacent regions can be exploited, and computation on overlapping regions is no longer required.

\begin{figure}[t] \small
	\centering
	\includegraphics[width=0.85\linewidth]{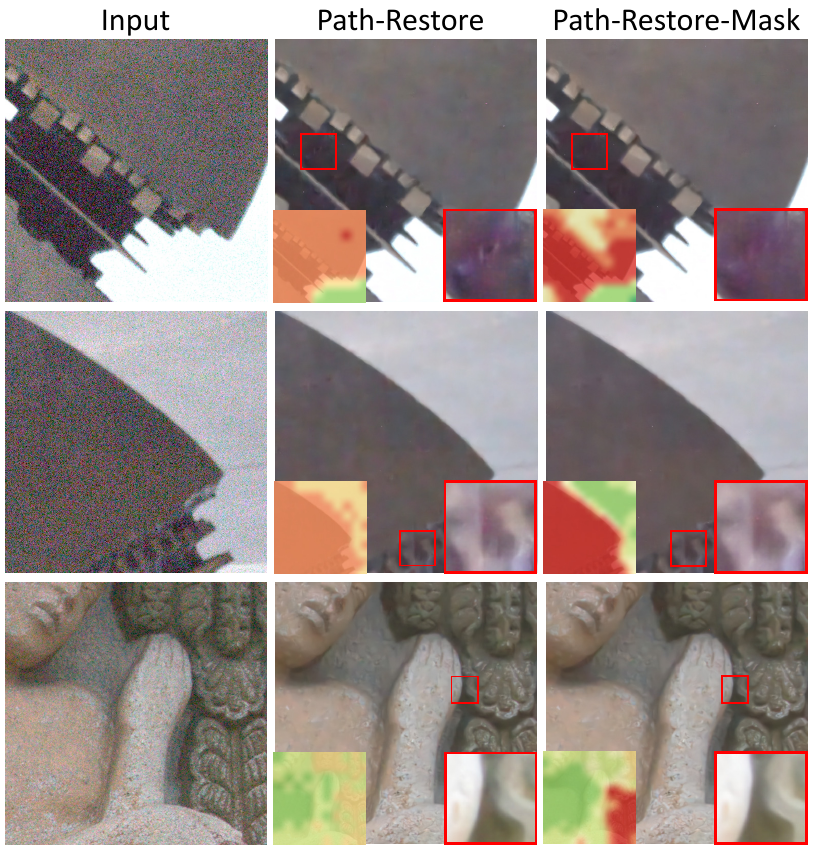}
	\caption{
		Qualitative comparisons between Path-Restore-Mask and Path-Restore. The brightness and contrast of the highlighted parts are increased for better view. The path-selection map is presented in the bottom left corner of each image.
	}
	\label{fig:image_CBD_mask}
	\vspace{-0.2cm}
\end{figure}

\noindent\textbf{Qualitative Results and Path Selection.}
Qualitative comparisons are shown in Figure~\ref{fig:image_CBD_mask}. We observe that most of the visual results are comparable, but Path-Restore-Mask yields fewer artifacts in the regions with severe noise (see the highlighted regions in the red box). The path-selection results indicate that Path-Restore-Mask selects more diverse paths with spatially smoother policy. Specifically, on the path-selection map of Path-Restore, the checkerboard-like policy occasionally appears, where one region is dispatched to a different path compared with that of surrounding regions. On the contrary, such a scenario seldom occurs for Path-Restore-Mask, because the pathfinder could observe the adjacent features while selecting the current path.

\begin{table}[t] \centering \small
	\center
	\caption{Ablation study on the number of paths in a dynamic block.}
	\begin{tabular}{c|c|c|c|c}
		\hline
		\multirow{2}{*}{Dataset} & \multicolumn{4}{c}{DIV2K-T50~\cite{agustsson2017ntire}} \\
		\cline{2-5}
		& \multicolumn{2}{c|}{63$\times$63 sub-image}         &  \multicolumn{2}{c}{2K image}    \\
		\hline
		Metric                   &   PSNR         & FLOPs (G)    & PSNR            & FLOPs (G) \\
		\hline\hline
		4 paths                  & \textbf{26.48} & \textbf{1.00}                      &  \textbf{25.81} & \textbf{1.38}  \\
		2 paths                  & 26.46          & 1.09                               &  25.80          & 1.48           \\
		\hline
		
	\end{tabular}
	\label{tab:ablation_architecture}
\end{table}

\begin{figure}[t] \small
	\centering
	\includegraphics[width=0.9\linewidth]{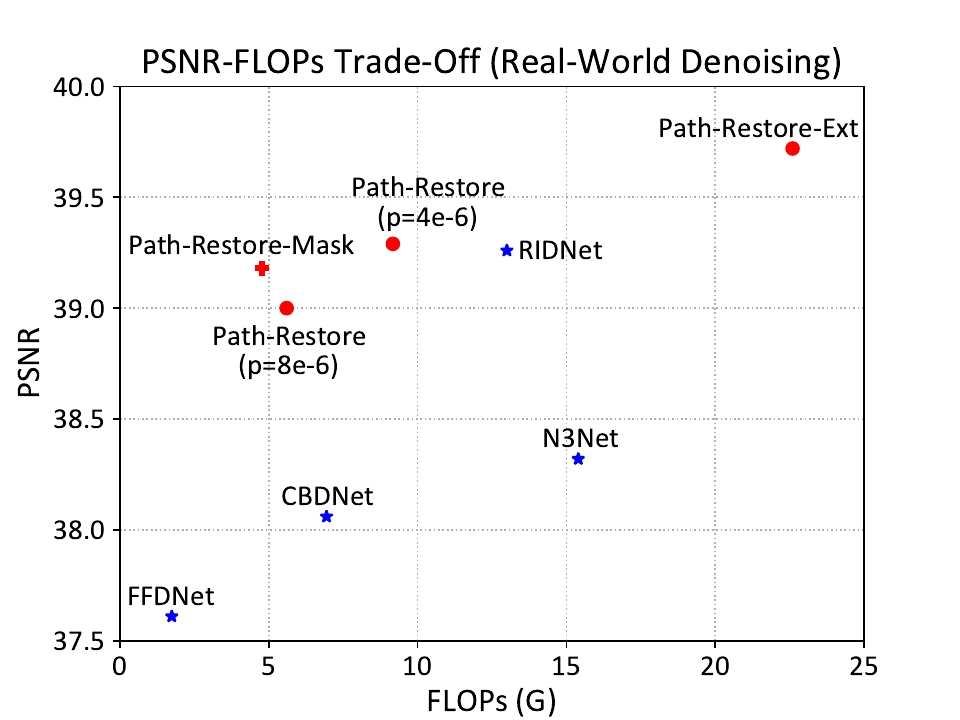}
	\caption{
		Performance-complexity trade-off for real-world denoising on the DND benchmark~\cite{plotz2017benchmarking}.
	}
	\label{fig:performance_complexity_real}
\end{figure}

\begin{figure}[t] \small
	\centering
	\includegraphics[width=0.9\linewidth]{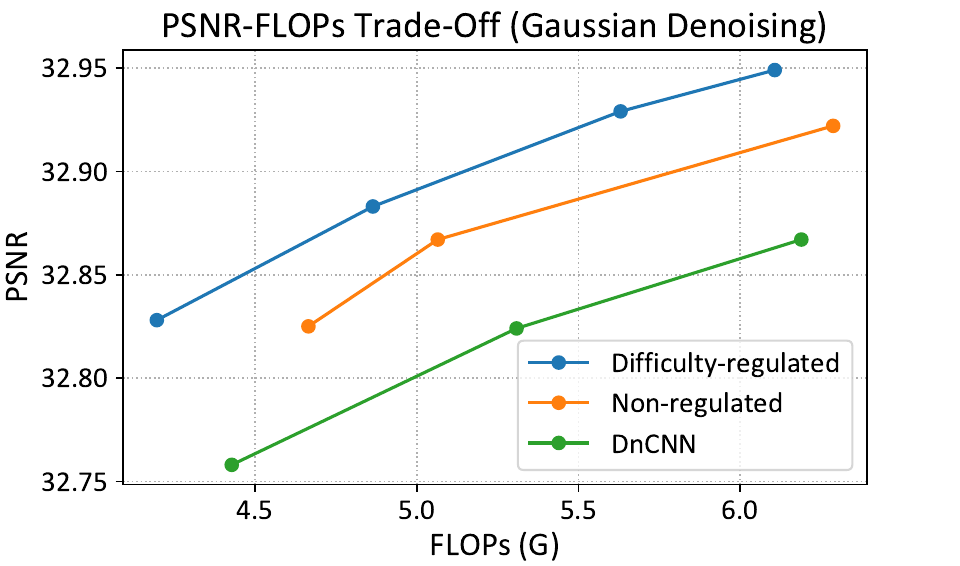}
	\caption{
		Performance-complexity trade-off vs. non-regulated reward and DnCNN for spatially variant (type ``linear'') Gaussian denoising on DIV2K-T50~\cite{agustsson2017ntire}.
	}
	\label{fig:performance_complexity}
\end{figure}

\subsection{Discussions}
\label{subsec:ablation}

\noindent\textbf{Architecture of Dynamic Block.}
We investigate the number of paths in each dynamic block. For the task to address mixed distortions, the number of paths is originally set as 4 since there are 3 paths for various distortions and 1 bypass connection. As shown in Table~\ref{tab:ablation_architecture}, we alternatively use 2 paths to address the same task. It is observed that the FLOPs increase by nearly 10\% when achieving comparable performance with the original setting. 

\noindent\textbf{Difficulty-Regulated Reward.}
A performance-complexity trade-off can be achieved by adjusting the reward penalty $p$ while training. The trade-off for real-world denoising on the DND benchmark~\cite{plotz2017benchmarking} is shown in Figure~\ref{fig:performance_complexity_real}, where the red points represent Path-Restore with different architecture and reward penalty while the blue points represent other methods. Path-Restore achieves a better PSNR-FLOPs trade-off compared with other methods when tuning the reward penalty. The trade-off is reasonable because a smaller reward penalty encourages the pathfinder to select a longer path for each region, resulting in better performance. We conduct a similar study on Gaussian denoising. As shown in Figure~\ref{fig:performance_complexity}, reducing $p$ gradually from $p=8\times10^{-6}$ to $3\times10^{-6}$, the blue curve depicts that PSNR and FLOPs both increase as the reward penalty decreases. We also adjust the depth of DnCNN to achieve the trade-off between PSNR and FLOPs, as shown in the green curve of Figure~\ref{fig:performance_complexity}. We observe that Path-Restore is consistently better than DnCNN by 0.1 dB with the same FLOPs. When achieving the same performance, our method is nearly 30\% faster than DnCNN.

\begin{figure}[t] \small
	\centering
	\includegraphics[width=0.8\linewidth]{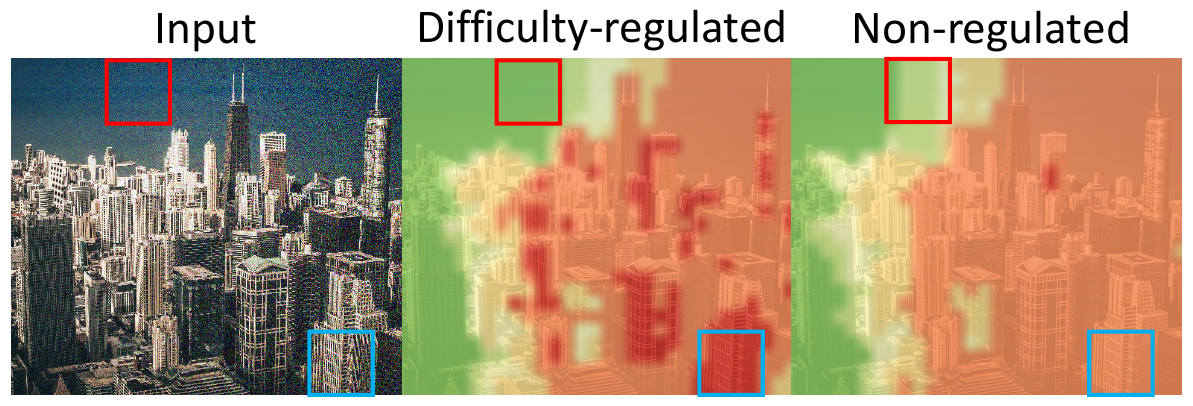}
	\caption{
		Path selection of regulated and non-regulated reward.
	}
	\label{fig:heatmap_rewards}
\end{figure}

We then study the impact of ``difficulty'' in the proposed reward function. In particular, we set difficulty $d\equiv1$ in Eq.~\eqref{eq:difficulty} to form a non-regulated reward. As shown in Figure~\ref{fig:performance_complexity}, the orange curve is consistently below the blue curve, indicating that the proposed difficulty-regulated reward helps Path-Restore achieve a better performance-complexity trade-off. We further present the path-selection policy of different rewards in Figure~\ref{fig:heatmap_rewards}. Compared with non-regulated reward, our difficulty-regulated reward saves more computations when processing easy regions (\eg, region in the red box), despite using a longer path to process hard regions (\eg, region in the blue box).

\begin{figure}[t] \small
	\centering
	\includegraphics[width=0.9\linewidth]{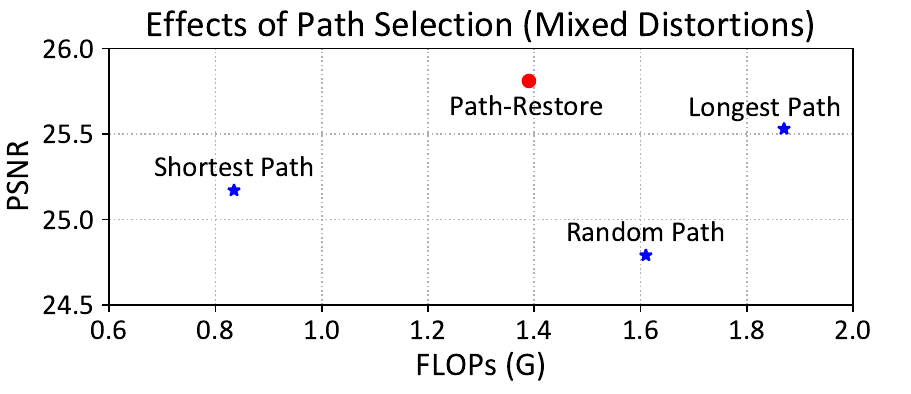}
	\caption{
		Effects of path selection for addressing mixed distortions.
	}
	\label{fig:path_selection_mixed_quantitative}
\end{figure}

\noindent\textbf{Training Datasets.}
The selection of training data is important for real-world denoising. In the aforementioned experiments, we adopt the same training data as CBDNet~\cite{guo2019toward} when evaluating our method on the DND~\cite{plotz2017benchmarking} benchmark, and only SIDD~\cite{SIDD_2018_CVPR} training dataset is used for the evaluation on SIDD benchmark. To investigate the effects of training datasets, we further use the same training data as RIDNet~\cite{anwar2019real} for the evaluation on both DND and SIDD. As shown in Table~\ref{tab:data_pathfinder}, Path-Restore-Mask represents our results with the original training data, and ``+ data of RIDNet'' denotes using the same training data as RIDNet. It is observed that the inference time almost remains the same. The PSNR is improved by 0.19 dB and 0.08 dB on DND and SIDD, respectively. The PSNR improvement is reasonable as the training dataset of RIDNet is larger than that of CBDNet.

\begin{table}[t]\centering \small
	\center
	\caption{Ablation study on training datasets and the pathfinder. ``+ data of RIDNet'' denotes using the same training data as RIDNet. ``- pathfinder'' also adopts the data of RIDNet, and the pathfinder is removed.}
	\setlength{\tabcolsep}{1.8mm}
	\begin{tabular}{c|c|c|c|c}
		\hline
		Dataset           & \multicolumn{2}{c|}{DND~\cite{plotz2017benchmarking}} & \multicolumn{2}{c}{SIDD~\cite{SIDD_2018_CVPR}} \\ \hline
		Metric            & PSNR        & Time (s)   & PSNR         & Time (s)   \\ \hline\hline
		Path-Restore-Mask & 39.18       & 0.146      & 38.42        & 0.76       \\
		+ data of RIDNet~\cite{anwar2019real}  & 39.37       & 0.143      & 38.50        & 0.74       \\
		- pathfinder      & 39.37       & 0.199      & 38.23        & 0.86       \\ \hline
	\end{tabular}
	\label{tab:data_pathfinder}
\end{table}

\begin{figure}[t] \small
	\centering
	\includegraphics[width=0.9\linewidth]{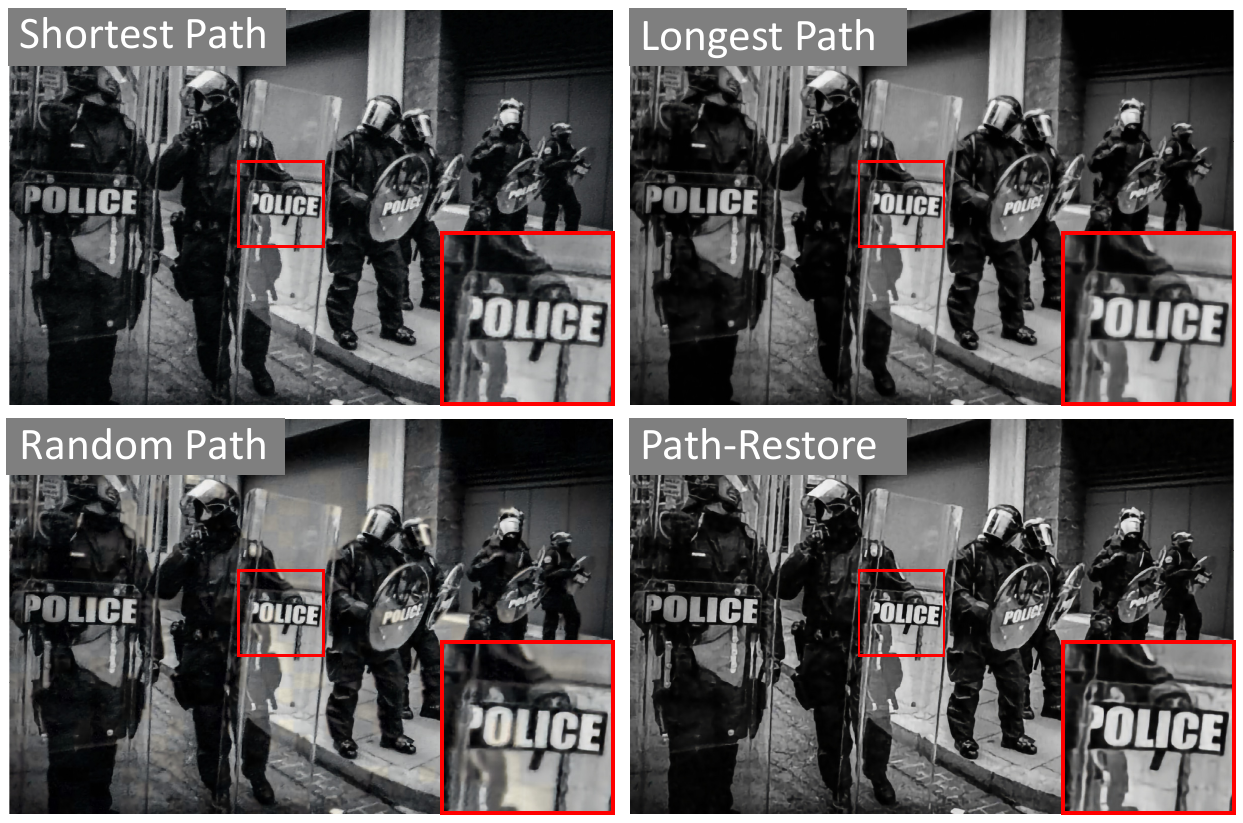}
	\caption{
		Qualitative comparisons of path selection for addressing mixed distortions (zoom in for best view).
	}
	\label{fig:path_selection_mixed_qualitative}
\end{figure}

\begin{table}[t]\centering \small
	\center
	\caption{Total parameters of each method on different tasks.}
	\begin{tabular}{c|c|c}
		\hline
		Task                      		    & Method       						& Parameters ($\times{10^5}$) \\ \hline\hline
		\multirow{5}{*}{Denoising}  		& FFDNet~\cite{zhang2018ffdnet} 	& 8.52                        \\
		& CBDNet~\cite{guo2019toward}   	& 67.9                        \\
		& N3Net~\cite{plotz2018neural}  	& 37.1                        \\
		& RIDNet~\cite{anwar2019real}   	& 16.4                        \\
		& PRIDNet~\cite{zhao2019pyramid}    & 742       				  \\
		& Path-Restore 						& 9.19                        \\ \hline
		\multirow{2}{*}{Mixed Distortions}	& RL-Restore~\cite{yu2018crafting}  & 5.22                        \\	
		& Path-Restore 						& 3.79                        \\ \hline
	\end{tabular}
	\label{tab:parameter}
\end{table}

\begin{figure}[t] \small
	\centering
	\includegraphics[width=0.92\linewidth]{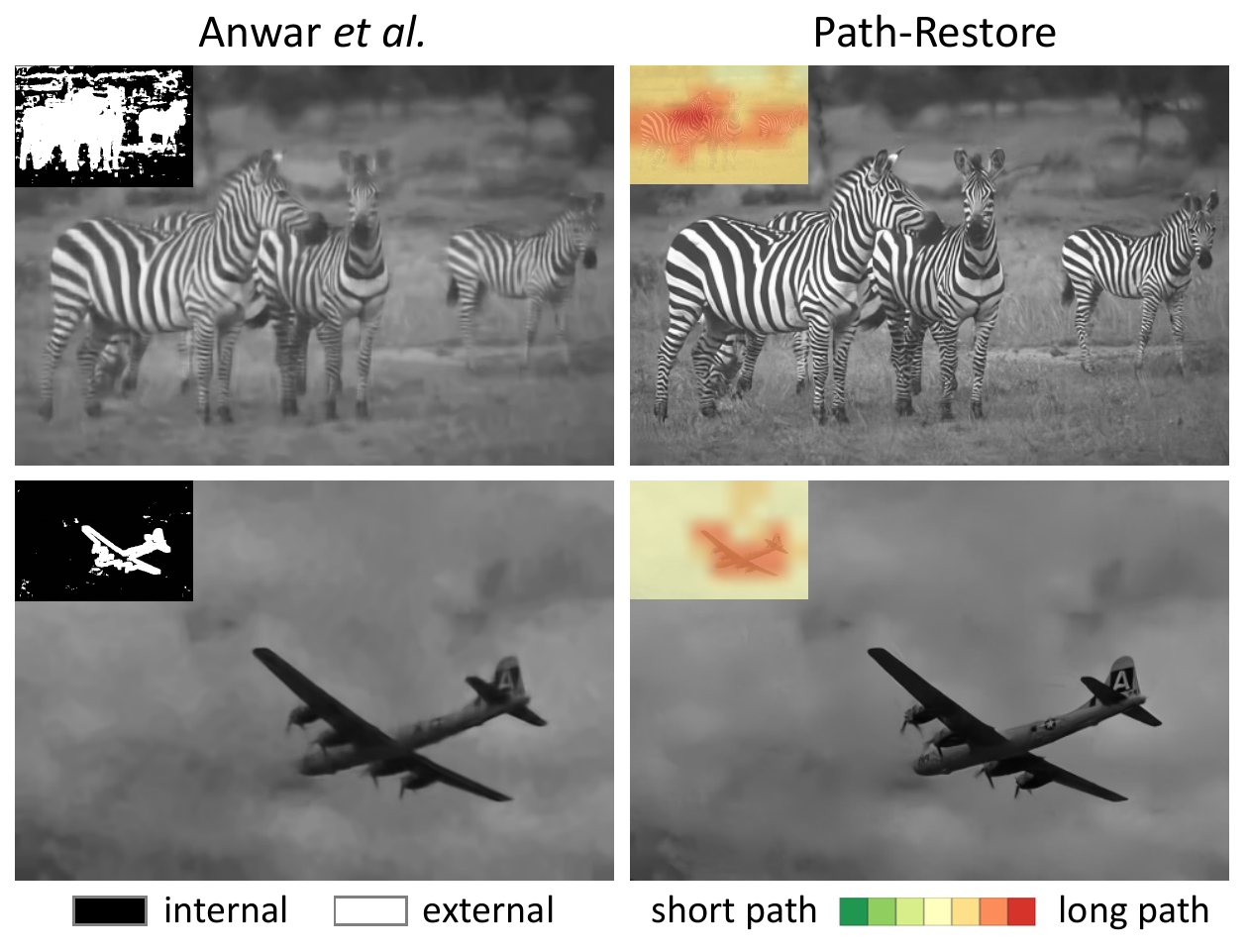}
	\vspace{-0.1cm}
	\caption{
		Qualitative comparisons to category-specific denoising~\cite{anwar2017combined}.
	}
	\label{fig:category_specific}
\end{figure}

\noindent\textbf{The Effects of Pathfinder.}
The pathfinder is an essential module in our framework. In Table~\ref{tab:data_pathfinder}, ``- pathfinder'' denotes a Path-Restore-Mask model without the pathfinder. In this case all image regions are processed by the same architecture. Note that we also use the training dataset of RIDNet~\cite{anwar2019real} to train this model. On DND~\cite{plotz2017benchmarking} benchmark, removing the pathfinder results in the same PSNR with an increase of 39\% in runtime. On SIDD~\cite{SIDD_2018_CVPR} benchmark, without the pathfinder, the PSNR decreases by 0.19 dB while the runtime increases by 13\%. 

We further evaluate the effects of pathfinder for addressing mixed distortions on DIV2K-T50 dataset~\cite{agustsson2017ntire}. We manually select the shortest path and one of the longest paths (the third path in each dynamic block), denoted by ``Shortest Path'' and ``Longest Path'', respectively. A ``Random path'' dispatch policy with uniform distribution is also used. As shown in Figure~\ref{fig:path_selection_mixed_quantitative}, with an effective pathfinder, Path-Restore outperforms all of the above policies. A random path selection is worse than the shortest path, indicating that learning path selection is non-trivial. Qualitative results are presented in Figure~\ref{fig:path_selection_mixed_qualitative}. Compared to manually designed dispatch policies, Path-Restore yields clearer images with fewer artifacts.

\noindent\textbf{Parameters.}
We summarize the number of parameters of each method in Table~\ref{tab:parameter}. The parameters of all paths are included. On the denoising task, Path-Restore has slightly more parameters than FFDNet, yet has apparently fewer parameters than other methods. Several approaches like CBDNet and PRIDNet adopt down-sampling for acceleration. However, it comes at a price of increasing parameters because the number of channels has to be expanded in a smaller scale. In particular, the number of parameters for PRIDNet is 140 times larger than ours. For addressing mixed distortions, Path-Restore has fewer parameters than RL-Restore. RL-Restore offers a toolbox that contains 12 task-specific CNNs. Although each CNN is light-weight, the total number of parameters is large.

\begin{table}[t] \centering\small
	\center
	\caption{Quantitative comparisons to category-specific denoising~\cite{anwar2017combined} on CBSD68 dataset with Gaussian noise $\sigma=30$.}
	\setlength{\tabcolsep}{4mm}
	\begin{tabular}{c|c|c|c}
		\hline
		Method                                       & PSNR  & SSIM   & Time (s) \\ \hline\hline
		Anwar~\etal~\cite{anwar2017combined}         & 25.06 & 0.6863 & 2.77$\times10^3$        \\ 
		Path-Restore                                 & 30.75 & 0.8742 & 2.67        \\ \hline
	\end{tabular}
	\label{tab:category_specific}
\end{table}

\noindent\textbf{Comparisons to Category-Specific Denoising.}
Anwar~\etal~\cite{anwar2017combined} propose a denoising algorithm that separately processes different image regions. Specifically, it applies internal denoising to smooth regions while conducts category-specific external denoising for textured regions. We evaluate this method on CBSD68 dataset with Gaussian noise level $\sigma=30$. The external dataset is the same as ours for Gaussian denoising.

Quantitative results are shown in Table~\ref{tab:category_specific}. The metrics are calculated on grayscale images. Our performance and efficiency significantly surpass that of~\cite{anwar2017combined} -- more than 5 dB higher in PSNR and 1,000 times faster. As Anwar~\etal~\cite{anwar2017combined} leverage PatchMatch algorithm~\cite{barnes2009patchmatch} for external denoising, the computational cost is very high, especially for high-resolution images. Although the category-specific method works well on images that contain regular patterns such as face and text, it is not as effective as learning-based methods for natural images where similar patches cannot be always found.

Qualitative results are presented in Figure~\ref{fig:category_specific}. Path-Restore produces clear and sharp results while category-specific denoising~\cite{anwar2017combined} yields blurry output. Interestingly, the denoising map of~\cite{anwar2017combined} is similar to our path-selection map. For category-specific denoising, the textured foreground is processed by external denoising while the smooth background is addressed by internal denoising. The criterion of whether a patch belongs to textured region is manually defined. As for Path-Restore, driven by a difficulty-regulated reward, the pathfinder learns to allocate textured and smooth regions to long and short paths, respectively. It demonstrates that the learned dispatch policy in our method conforms to human prior.

\section{Conclusion}
We have devised a new framework for image restoration with low computational cost. Combining reinforcement learning and deep learning, we propose Path-Restore that enables path selection for each image region. Specifically, Path-Restore is composed of a multi-path CNN and a pathfinder. The multi-path CNN offers several paths to address different types of distortions, and the pathfinder is able to find the optimal path to efficiently restore each region. To learn a reasonable dispatch policy, we propose a difficulty-regulated reward that encourages the pathfinder to select short paths for the regions that are easy to restore. We further investigate two mechanisms to independently and jointly process different image regions. Path-Restore achieves comparable or superior performance to existing methods on various restoration tasks with much fewer computations. Our method is especially effective for real-world denoising, where the noise distribution is diverse across different image regions.
The idea of region-based path selection can also be applied to other tasks where each image region should be treated differently.

\section*{Acknowledgment}
This work is partially supported by the Collaborative Research grant from SenseTime Group (CUHK Agreement No. TS1610626 \& No. TS1712093), Singapore MOE AcRF Tier 1 (2018-T1-002-056), NTU NAP. This study is supported under the RIE2020 Industry Alignment Fund – Industry Collaboration Projects (IAF-ICP) Funding Initiative, as well as cash and in-kind contribution from the industry partner(s). It is also supported in part by the National Natural Science Foundation of China (61906184), the Joint Lab of CAS-HK.

{\small
	\bibliographystyle{IEEEtran}
	\bibliography{short,routingIR_pami_v2}

\begin{thebibliography}{10}
\providecommand{\url}[1]{#1}
\csname url@samestyle\endcsname
\providecommand{\newblock}{\relax}
\providecommand{\bibinfo}[2]{#2}
\providecommand{\BIBentrySTDinterwordspacing}{\spaceskip=0pt\relax}
\providecommand{\BIBentryALTinterwordstretchfactor}{4}
\providecommand{\BIBentryALTinterwordspacing}{\spaceskip=\fontdimen2\font plus
\BIBentryALTinterwordstretchfactor\fontdimen3\font minus
  \fontdimen4\font\relax}
\providecommand{\BIBforeignlanguage}[2]{{%
\expandafter\ifx\csname l@#1\endcsname\relax
\typeout{** WARNING: IEEEtran.bst: No hyphenation pattern has been}%
\typeout{** loaded for the language `#1'. Using the pattern for}%
\typeout{** the default language instead.}%
\else
\language=\csname l@#1\endcsname
\fi
#2}}
\providecommand{\BIBdecl}{\relax}
\BIBdecl

\bibitem{anwar2019real}
S.~Anwar and N.~Barnes, ``Real image denoising with feature attention,'' in
  \emph{Proceedings of the IEEE International Conference on Computer Vision},
  2019, pp. 3155--3164.

\bibitem{plotz2017benchmarking}
T.~Plotz and S.~Roth, ``Benchmarking denoising algorithms with real
  photographs,'' in \emph{Proceedings of the IEEE Conference on Computer Vision
  and Pattern Recognition}, 2017, pp. 1586--1595.

\bibitem{timofte2018ntire}
R.~Timofte, S.~Gu, J.~Wu, L.~V. Gool, L.~Zhang \emph{et~al.}, ``Ntire 2018
  challenge on single image super-resolution: Methods and results,'' in
  \emph{Proceedings of the IEEE Conference on Computer Vision and Pattern
  Recognition Workshops}, 2018, pp. 852--863.

\bibitem{lim2017enhanced}
B.~Lim, S.~Son, H.~Kim, S.~Nah, and K.~M. Lee, ``Enhanced deep residual
  networks for single image super-resolution,'' in \emph{Proceedings of the
  IEEE Conference on Computer Vision and Pattern Recognition Workshops}, 2017,
  pp. 136--144.

\bibitem{huang2017densely}
G.~Huang, Z.~Liu, L.~Van Der~Maaten, and K.~Q. Weinberger, ``Densely connected
  convolutional networks.'' in \emph{Proceedings of the IEEE Conference on
  Computer Vision and Pattern Recognition}, 2017, pp. 4700--4708.

\bibitem{lefkimmiatis2016non}
S.~Lefkimmiatis, ``Non-local color image denoising with convolutional neural
  networks,'' in \emph{Proceedings of the IEEE Conference on Computer Vision
  and Pattern Recognition}, 2017, pp. 3587--3596.

\bibitem{jain2009natural}
V.~Jain and S.~Seung, ``Natural image denoising with convolutional networks,''
  in \emph{Proceedings of the Advances in Neural Information Processing
  Systems}, 2009, pp. 769--776.

\bibitem{chen2018image}
J.~Chen, J.~Chen, H.~Chao, and M.~Yang, ``Image blind denoising with generative
  adversarial network based noise modeling,'' in \emph{Proceedings of the IEEE
  Conference on Computer Vision and Pattern Recognition}, 2018, pp. 3155--3164.

\bibitem{guo2019toward}
S.~Guo, Z.~Yan, K.~Zhang, W.~Zuo, and L.~Zhang, ``Toward convolutional blind
  denoising of real photographs,'' in \emph{Proceedings of the IEEE Conference
  on Computer Vision and Pattern Recognition}, 2019, pp. 1712--1722.

\bibitem{xu2014Deep}
L.~Xu, J.~S.~J. Ren, C.~Liu, and J.~Jia, ``Deep convolutional neural network
  for image deconvolution,'' in \emph{Proceedings of the Advances in Neural
  Information Processing Systems}, 2014, pp. 1790--1798.

\bibitem{sun2015learning}
J.~Sun, W.~Cao, Z.~Xu, and J.~Ponce, ``Learning a convolutional neural network
  for non-uniform motion blur removal,'' in \emph{Proceedings of the IEEE
  Conference on Computer Vision and Pattern Recognition}, 2015, pp. 769--777.

\bibitem{nah2017deep}
S.~Nah, T.~H. Kim, and K.~M. Lee, ``Deep multi-scale convolutional neural
  network for dynamic scene deblurring,'' in \emph{Proceedings of the IEEE
  Conference on Computer Vision and Pattern Recognition}, 2017, pp. 257--265.

\bibitem{dong2015compression}
C.~Dong, Y.~Deng, C.~Change~Loy, and X.~Tang, ``Compression artifacts reduction
  by a deep convolutional network,'' in \emph{Proceedings of the IEEE
  International Conference on Computer Vision}, 2015, pp. 576--584.

\bibitem{wang2016d3}
Z.~Wang, D.~Liu, S.~Chang, Q.~Ling, Y.~Yang, and T.~S. Huang, ``D3: Deep
  dual-domain based fast restoration of {JPEG}-compressed images,'' in
  \emph{Proceedings of the IEEE Conference on Computer Vision and Pattern
  Recognition}, 2016, pp. 2764--2772.

\bibitem{guo2016building}
J.~Guo and H.~Chao, ``Building dual-domain representations for compression
  artifacts reduction,'' in \emph{Proceedings of the European Conference on
  Computer Vision}, 2016, pp. 628–--644.

\bibitem{guo2016one}
------, ``One-to-many network for visually pleasing compression artifacts
  reduction,'' in \emph{Proceedings of the IEEE Conference on Computer Vision
  and Pattern Recognition}, 2017, pp. 3038--3047.

\bibitem{dong2016image}
C.~Dong, C.~C. Loy, K.~He, and X.~Tang, ``Image super-resolution using deep
  convolutional networks,'' \emph{IEEE Transactions on Pattern Analysis and
  Machine Intelligence}, vol.~38, no.~2, pp. 295--307, 2016.

\bibitem{kim2016accurate}
J.~Kim, J.~Kwon~Lee, and K.~Mu~Lee, ``Accurate image super-resolution using
  very deep convolutional networks,'' in \emph{Proceedings of the IEEE
  Conference on Computer Vision and Pattern Recognition}, 2016, pp.
  1646–--1654.

\bibitem{kim2016deeply}
J.~Kim, J.~K. Lee, and K.~M. Lee, ``Deeply-recursive convolutional network for
  image super-resolution,'' in \emph{Proceedings of the IEEE Conference on
  Computer Vision and Pattern Recognition}, 2016, pp. 1637–--1645.

\bibitem{tai2017image}
Y.~Tai, J.~Yang, and X.~Liu, ``Image super-resolution via deep recursive
  residual network,'' in \emph{Proceedings of the IEEE Conference on Computer
  Vision and Pattern Recognition}, 2017, pp. 2790–--2798.

\bibitem{ledig2017photo}
C.~Ledig, L.~Theis, F.~Husz{\'a}r, J.~Caballero, A.~Cunningham, A.~Acosta,
  A.~Aitken, A.~Tejani, J.~Totz, Z.~Wang \emph{et~al.}, ``Photo-realistic
  single image super-resolution using a generative adversarial network,'' in
  \emph{Proceedings of the IEEE Conference on Computer Vision and Pattern
  Recognition}, 2017, pp. 105--114.

\bibitem{wang2018esrgan}
X.~Wang, K.~Yu, S.~Wu, J.~Gu, Y.~Liu, C.~Dong, Y.~Qiao, and C.~Change~Loy,
  ``Esrgan: Enhanced super-resolution generative adversarial networks,'' in
  \emph{Proceedings of the European Conference on Computer Vision Workshops},
  2018, pp. 63--79.

\bibitem{zhang2017beyond}
K.~Zhang, W.~Zuo, Y.~Chen, D.~Meng, and L.~Zhang, ``Beyond a gaussian denoiser:
  Residual learning of deep cnn for image denoising,'' \emph{IEEE Transactions
  on Image Processing}, vol.~26, no.~7, pp. 3142--3155, 2017.

\bibitem{plotz2018neural}
T.~Pl{\"o}tz and S.~Roth, ``Neural nearest neighbors networks,'' in
  \emph{Proceedings of the Advances in Neural Information Processing Systems},
  2018, pp. 1087--1098.

\bibitem{zhao2019pyramid}
Y.~Zhao, Z.~Jiang, A.~Men, and G.~Ju, ``Pyramid real image denoising network,''
  in \emph{Proceedings of the IEEE Visual Communications and Image Processing},
  2019, pp. 1--4.

\bibitem{anwar2017combined}
S.~Anwar, C.~P. Huynh, and F.~Porikli, ``Combined internal and external
  category-specific image denoising.'' in \emph{Proceedings of the British
  Machine Vision Conference}, 2017, pp. 71.1--71.12.

\bibitem{yu2018crafting}
K.~Yu, C.~Dong, L.~Lin, and C.~C. Loy, ``Crafting a toolchain for image
  restoration by deep reinforcement learning,'' in \emph{Proceedings of the
  IEEE Conference on Computer Vision and Pattern Recognition}, 2018, pp.
  2443--2452.

\bibitem{bengio2013estimating}
Y.~Bengio, N.~L{\'e}onard, and A.~Courville, ``Estimating or propagating
  gradients through stochastic neurons for conditional computation,''
  \emph{arXiv preprint arXiv:1308.3432}, 2013.

\bibitem{figurnov2017spatially}
M.~Figurnov, M.~D. Collins, Y.~Zhu, L.~Zhang, J.~Huang, D.~Vetrov, and
  R.~Salakhutdinov, ``Spatially adaptive computation time for residual
  networks,'' in \emph{Proceedings of the IEEE Conference on Computer Vision
  and Pattern Recognition}, 2017, pp. 1039--1048.

\bibitem{wu2018blockdrop}
Z.~Wu, T.~Nagarajan, A.~Kumar, S.~Rennie, L.~S. Davis, K.~Grauman, and
  R.~Feris, ``Blockdrop: Dynamic inference paths in residual networks,'' in
  \emph{Proceedings of the IEEE Conference on Computer Vision and Pattern
  Recognition}, 2018, pp. 8817--8826.

\bibitem{wang2017skipnet}
X.~Wang, F.~Yu, Z.-Y. Dou, and J.~E. Gonzalez, ``Skipnet: Learning dynamic
  routing in convolutional networks,'' in \emph{Proceedings of the European
  Conference on Computer Vision}, 2018, pp. 409--424.

\bibitem{he2016deep}
K.~He, X.~Zhang, S.~Ren, and J.~Sun, ``Deep residual learning for image
  recognition,'' in \emph{Proceedings of the IEEE Conference on Computer Vision
  and Pattern Recognition}, 2016, pp. 770--778.

\bibitem{rosenbaum2017routing}
C.~Rosenbaum, T.~Klinger, and M.~Riemer, ``Routing networks: Adaptive selection
  of non-linear functions for multi-task learning,'' in \emph{Proceedings of
  the International Conference on Learning Representations}, 2018.

\bibitem{mnih2015human}
V.~Mnih, K.~Kavukcuoglu, D.~Silver, A.~A. Rusu, J.~Veness \emph{et~al.},
  ``Human-level control through deep reinforcement learning,'' \emph{Nature},
  vol. 518, no. 7540, pp. 529--533, 2015.

\bibitem{hessel2017rainbow}
M.~Hessel, J.~Modayil, H.~Van~Hasselt, T.~Schaul, G.~Ostrovski, W.~Dabney,
  D.~Horgan, B.~Piot, M.~Azar, and D.~Silver, ``Rainbow: Combining improvements
  in deep reinforcement learning,'' in \emph{Proceedings of the AAAI Conference
  on Artificial Intelligence}, 2018, pp. 3215--3222.

\bibitem{silver2017mastering}
D.~Silver, J.~Schrittwieser, K.~Simonyan, I.~Antonoglou, A.~Huang
  \emph{et~al.}, ``Mastering the game of {Go} without human knowledge,''
  \emph{Nature}, vol. 550, no. 7676, pp. 354--359, 2017.

\bibitem{baker2016designing}
B.~Baker, O.~Gupta, N.~Naik, and R.~Raskar, ``Designing neural network
  architectures using reinforcement learning,'' in \emph{Proceedings of the
  International Conference on Learning Representations}, 2017.

\bibitem{zoph2016neural}
B.~Zoph and Q.~V. Le, ``Neural architecture search with reinforcement
  learning,'' in \emph{Proceedings of the International Conference on Learning
  Representations}, 2017.

\bibitem{zhong2018practical}
Z.~Zhong, J.~Yan, W.~Wu, J.~Shao, and C.-L. Liu, ``Practical block-wise neural
  network architecture generation,'' in \emph{Proceedings of the IEEE
  Conference on Computer Vision and Pattern Recognition}, 2018, pp. 2423--2432.

\bibitem{hu2018exposure}
Y.~Hu, H.~He, C.~Xu, B.~Wang, and S.~Lin, ``Exposure: A white-box photo
  post-processing framework,'' \emph{ACM Transactions on Graphics}, vol.~37,
  no.~2, pp. 1--17, 2018.

\bibitem{jang2017categorical}
E.~Jang, S.~Gu, and B.~Poole, ``Categorical reparameterization with
  gumbel-softmax,'' in \emph{Proceedings of the International Conference on
  Learning Representations}, 2017.

\bibitem{veit2018convolutional}
A.~Veit and S.~Belongie, ``Convolutional networks with adaptive inference
  graphs,'' in \emph{Proceedings of the European Conference on Computer
  Vision}, 2018, pp. 3--18.

\bibitem{xie2019snas}
S.~Xie, H.~Zheng, C.~Liu, and L.~Lin, ``Snas: stochastic neural architecture
  search,'' in \emph{Proceedings of the International Conference on Learning
  Representations}, 2019.

\bibitem{wu2019fbnet}
B.~Wu, X.~Dai, P.~Zhang, Y.~Wang, F.~Sun, Y.~Wu, Y.~Tian, P.~Vajda, Y.~Jia, and
  K.~Keutzer, ``Fbnet: Hardware-aware efficient convnet design via
  differentiable neural architecture search,'' in \emph{Proceedings of the IEEE
  Conference on Computer Vision and Pattern Recognition}, 2019, pp.
  10\,726–--10\,734.

\bibitem{xingjian2015convolutional}
X.~Shi, Z.~Chen, H.~Wang, D.-Y. Yeung, W.-K. Wong, and W.-c. Woo,
  ``Convolutional lstm network: A machine learning approach for precipitation
  nowcasting,'' in \emph{Proceedings of the Advances in Neural Information
  Processing Systems}, 2015, pp. 802--810.

\bibitem{johnson2016perceptual}
J.~Johnson, A.~Alahi, and L.~Fei-Fei, ``Perceptual losses for real-time style
  transfer and super-resolution,'' in \emph{Proceedings of the European
  Conference on Computer Vision}, 2016, pp. 694--711.

\bibitem{williams1992simple}
R.~J. Williams, ``Simple statistical gradient-following algorithms for
  connectionist reinforcement learning,'' \emph{Machine Learning}, vol.~8, no.
  3-4, pp. 229--256, 1992.

\bibitem{kingma2015adam}
D.~P. Kingma and J.~Ba, ``Adam: A method for stochastic optimization,'' in
  \emph{Proceedings of the International Conference on Learning
  Representations}, 2015.

\bibitem{abadi2016tensorflow}
M.~Abadi, P.~Barham, J.~Chen, Z.~Chen, A.~Davis, J.~Dean, M.~Devin,
  S.~Ghemawat, G.~Irving, M.~Isard \emph{et~al.}, ``Tensorflow: A system for
  large-scale machine learning,'' in \emph{12th $\{$USENIX$\}$ Symposium on
  Operating Systems Design and Implementation ($\{$OSDI$\}$ 16)}, 2016, pp.
  265--283.

\bibitem{SIDD_2018_CVPR}
A.~Abdelhamed, S.~Lin, and M.~S. Brown, ``A high-quality denoising dataset for
  smartphone cameras,'' in \emph{Proceedings of the IEEE Conference on Computer
  Vision and Pattern Recognition}, 2018, pp. 1692--1700.

\bibitem{martin2001database}
D.~Martin, C.~Fowlkes, D.~Tal, and J.~Malik, ``A database of human segmented
  natural images and its application to evaluating segmentation algorithms and
  measuring ecological statistics,'' in \emph{Proceedings of the IEEE
  International Conference on Computer Vision}, 2001, pp. 416--423.

\bibitem{ma2017waterloo}
K.~Ma, Z.~Duanmu, Q.~Wu, Z.~Wang, H.~Yong, H.~Li, and L.~Zhang, ``Waterloo
  exploration database: New challenges for image quality assessment models,''
  \emph{IEEE Transactions on Image Processing}, vol.~26, no.~2, pp. 1004--1016,
  2017.

\bibitem{anaya2018renoir}
J.~Anaya and A.~Barbu, ``Renoir--a dataset for real low-light image noise
  reduction,'' \emph{Journal of Visual Communication and Image Representation},
  vol.~51, pp. 144--154, 2018.

\bibitem{dabov2007color}
K.~Dabov, A.~Foi, V.~Katkovnik, and K.~O. Egiazarian, ``Color image denoising
  via sparse 3d collaborative filtering with grouping constraint in
  luminance-chrominance space.'' in \emph{Proceedings of the IEEE International
  Conference on Image Processing}, 2007, pp. 313--316.

\bibitem{zhang2018ffdnet}
K.~Zhang, W.~Zuo, and L.~Zhang, ``Ffdnet: Toward a fast and flexible solution
  for cnn based image denoising,'' \emph{IEEE Transactions on Image
  Processing}, vol.~27, no.~9, pp. 4608--4622, 2018.

\bibitem{tseng2019hyperparameter}
E.~Tseng, F.~Yu, Y.~Yang, F.~Mannan, K.~S. Arnaud, D.~Nowrouzezahrai, J.-F.
  Lalonde, and F.~Heide, ``Hyperparameter optimization in black-box image
  processing using differentiable proxies,'' \emph{ACM Transactions on
  Graphics}, vol.~38, no.~4, p.~27, 2019.

\bibitem{roth2009fields}
S.~Roth and M.~J. Black, ``Fields of experts,'' \emph{International Journal of
  Computer Vision}, vol.~82, no.~2, p. 205, 2009.

\bibitem{agustsson2017ntire}
E.~Agustsson and R.~Timofte, ``Ntire 2017 challenge on single image
  super-resolution: Dataset and study,'' in \emph{Proceedings of the IEEE
  Conference on Computer Vision and Pattern Recognition Workshops}, 2017, pp.
  126--135.

\bibitem{brooks2018unprocessing}
T.~Brooks, B.~Mildenhall, T.~Xue, J.~Chen, D.~Sharlet, and J.~T. Barron,
  ``Unprocessing images for learned raw denoising,'' in \emph{Proceedings of
  the IEEE Conference on Computer Vision and Pattern Recognition}, 2019, pp.
  11\,036--11\,045.

\bibitem{chen2017trainable}
Y.~Chen and T.~Pock, ``Trainable nonlinear reaction diffusion: A flexible
  framework for fast and effective image restoration,'' \emph{IEEE Transactions
  on Pattern Analysis and Machine Intelligence}, vol.~39, no.~6, pp.
  1256--1272, 2017.

\bibitem{barnes2009patchmatch}
C.~Barnes, E.~Shechtman, A.~Finkelstein, and D.~B. Goldman, ``Patchmatch: A
  randomized correspondence algorithm for structural image editing,'' \emph{ACM
  Transactions on Graphics}, vol.~28, no.~3, p.~24, 2009.

\end{thebibliography}
}

%

\begin{IEEEbiography}[{\includegraphics[width=1.0in,height=1.25in,clip,keepaspectratio]{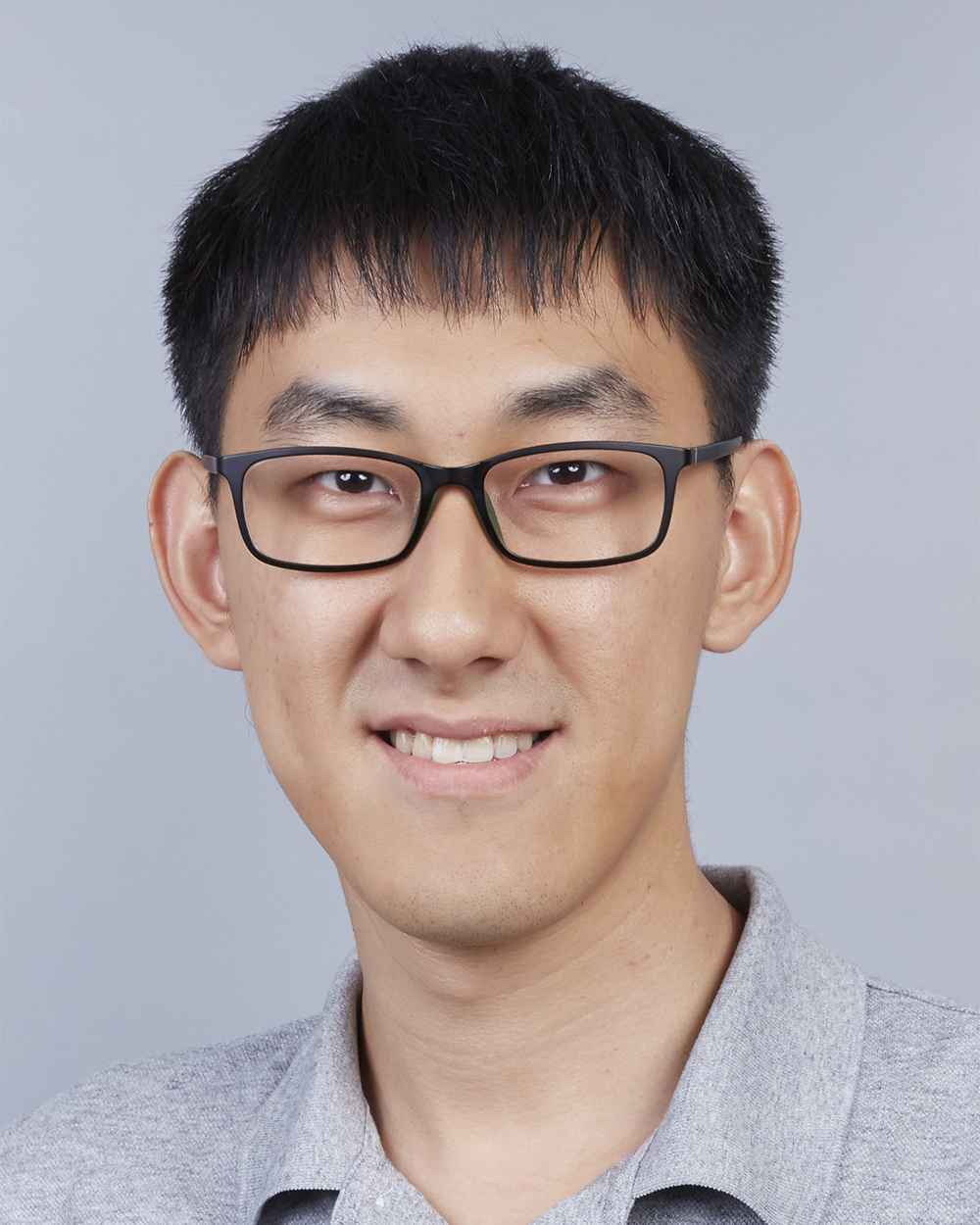}}]{Ke Yu}
	received a B.Eng. degree at the Department of Electronic Engineering from Tsinghua University, in 2016. He is currently pursuing a Ph.D. degree in the Department of Information Engineering, The Chinese University of Hong Kong. He was selected as an outstanding reviewer in CVPR 2019. He won the first place in several international super-resolution challenges including NTIRE2019 and PIRM2018. His research interests include image restoration, super-resolution and reinforcement learning.
\end{IEEEbiography}

\begin{IEEEbiography}[{\includegraphics[width=1.0in,height=1.25in,clip,keepaspectratio]{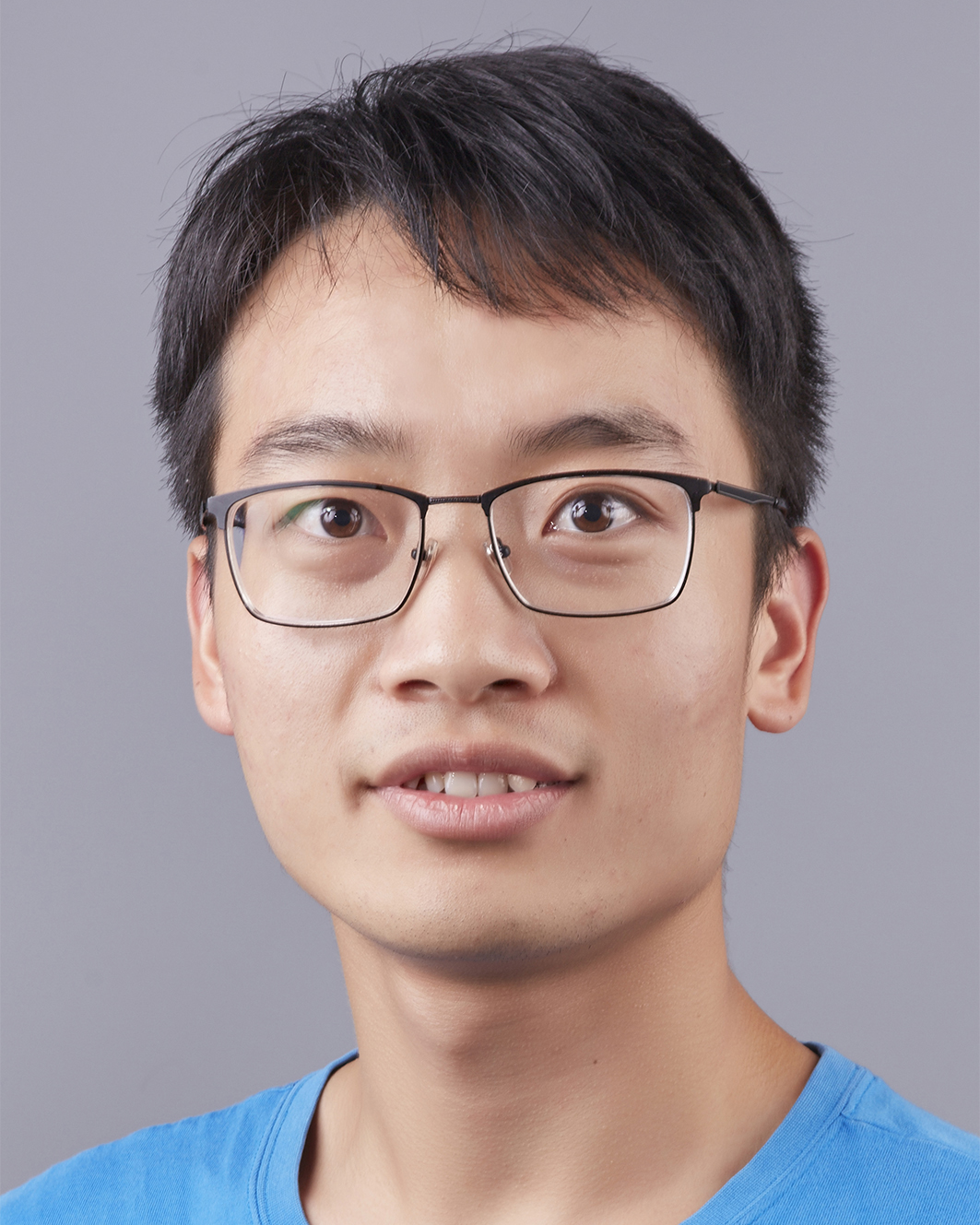}}]{Xintao Wang}
	is currently a researcher in Applied Research Center (ARC), Tencent PCG. 
	He received his Ph.D. degree in the Department of Information Engineering, The Chinese University of Hong Kong, in 2020.
	He was selected as an outstanding reviewer in CVPR 2019 and an outstanding reviewer (honorable mention) in BMVC 2019. 
	He won the first place in several international super-resolution challenges including NTIRE2019, NTIRE2018, and PIRM2018.
	His research interests focus on low-level vision problems, including super-resolution, image and video restoration.
\end{IEEEbiography}

\begin{IEEEbiography}[{\includegraphics[width=1.0in,height=1.25in,clip,keepaspectratio]{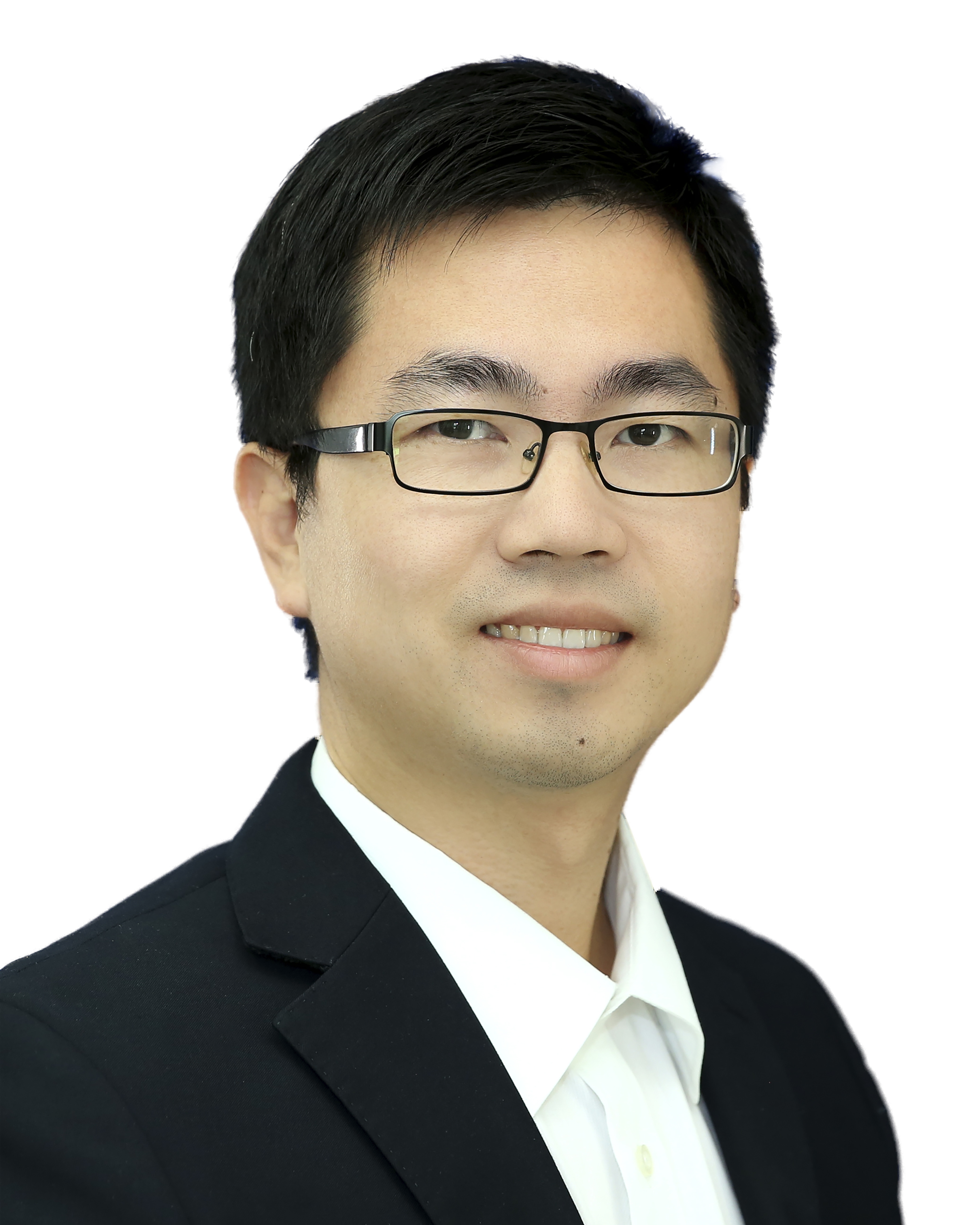}}]{Chao Dong}
	is currently an associate professor in Shenzhen Institute of Advanced Technology, Chinese Academy of Science. He received his Ph.D. degree from The Chinese University of Hong Kong in 2016. In 2014, he first introduced deep learning method -- SRCNN into the super-resolution field. This seminal work was chosen as one of the top ten “Most Popular Articles” of TPAMI in 2016. His team has won several championships in international challenges – NTIRE2018, PIRM2018, NTIRE2019, NTIRE2020 and AIM2020. He worked in SenseTime from 2016 to 2018, as the team leader of Super-Resolution Group. His current research interest focuses on low-level vision problems, such as image/video super-resolution, denoising and enhancement.
\end{IEEEbiography}

\begin{IEEEbiography}[{\includegraphics[width=1.0in,height=1.25in,clip,keepaspectratio]{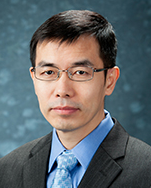}}]{Xiaoou Tang}
	(S’93-M’96-SM’02-F’09) received a B.S. degree from the University of Science and Technology of China, Hefei, in 1990, a M.S. degree from the University of Rochester, New York, in 1991, and a Ph.D. degree from the Massachusetts Institute of Technology, Cambridge, in 1996. He is a Professor of the Department of Information Engineering, The Chinese University of Hong Kong. He worked as a group manager of the Visual Computing Group at the Microsoft Research Asia, from 2005 to 2008. His research interests include computer vision, pattern recognition and video processing. He received the Best Paper Award at the IEEE Conference on Computer Vision and Pattern Recognition (CVPR) 2009 and Outstanding Student Paper Award at the AAAI 2015. He was a program chair of the IEEE International Conference on Computer Vision (ICCV) 2009, a General Chair of ICCV 2019, and served as an associate editor of the IEEE Transactions on Pattern Analysis and Machine Intelligence and Editor-in-Chief of the International Journal of Computer Vision. He is a fellow of the IEEE. 
\end{IEEEbiography}

\begin{IEEEbiography}[{\includegraphics[width=1.0in,height=1.25in,clip,keepaspectratio]{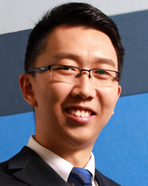}}]{Chen Change Loy}
	is an Associate Professor with the School of Computer Science and Engineering, Nanyang Technological University, Singapore. He is also an Adjunct Associate Professor at The Chinese University of Hong Kong. He received his Ph.D. (2010) in Computer Science from the Queen Mary University of London. Prior to joining NTU, he served as a Research Assistant Professor at the MMLab of The Chinese University of Hong Kong, from 2013 to 2018. He was a postdoctoral researcher at Queen Mary University of London and Vision Semantics Limited, from 2010 to 2013.
	He serves as an Associate Editor of the IEEE Transactions on Pattern Analysis and Machine Intelligence and   International Journal of Computer Vision. He also serves/served as an Area Chair of major conferences such as ICCV 2021, CVPR 2021, CVPR 2019, and ECCV 2018. He is a senior member of IEEE.
	His research interests include image/video restoration and enhancement, generative tasks, and representation learning.
\end{IEEEbiography}







\end{document}